\documentclass[10pt,twocolumn,letterpaper]{article}

\usepackage{cvpr}
\usepackage{times}
\usepackage{epsfig}
\usepackage{graphicx}
\usepackage{amsmath}
\usepackage{amssymb}
\usepackage{subfigure}
\usepackage{multirow} % Cells that span multiple rows in a table
\usepackage{multicol}

\newcommand{\mypar}[1]{\vspace{-6mm}\paragraph{#1}} % Customized paragraph without vspace
\usepackage[normalem]{ulem} % for striking out text with \sout
\newcommand{\figref}[1]{Figure~\ref{#1}}
\newcommand{\tableref}[1]{Table~\ref{#1}}
\newcommand{\secref}[1]{Section~\ref{#1}}
\newcommand{\squeeze}{\vspace{-1mm}} % To add in various places to quickly adjust whitespaces
\usepackage{tabularx}
\newcolumntype{C}[1]{>{\centering\let\newline\\\arraybackslash\hspace{0pt}}m{#1}} % column-type in table
\newcolumntype{L}[1]{>{\let\newline\\\arraybackslash\hspace{0pt}}m{#1}} % column-type in table
\usepackage{siunitx} % For degrees symbol
\usepackage{float} % For figure placement

\usepackage[usenames, dvipsnames]{color} % for custom colors
\definecolor{purple}{rgb}{0.5, 0.0, 0.5}
\definecolor{orange}{rgb}{1, 0.65, 0}
\definecolor{lightgreen}{rgb}{0.68, 1, 0.18}
\definecolor{darkgreen}{rgb}{0.09, 0.32, 0.24}
\definecolor{darkred}{rgb}{0.6, 0, 0}
\definecolor{brown}{rgb}{0.64, 0.16, 0.16}
\definecolor{lilac}{rgb}{0.82, 0.67, 1}

\usepackage[pagebackref=true,breaklinks=true,letterpaper=true,colorlinks,bookmarks=false]{hyperref}

\cvprfinalcopy

\ifcvprfinal\fi
\begin{document}

%% Limit space around equations that use "align". Must be after \begin{document}.
\setlength{\abovedisplayskip}{5pt}
\setlength{\belowdisplayskip}{5pt}
\setlength{\abovedisplayshortskip}{0pt}
\setlength{\belowdisplayshortskip}{0pt}

%%%%%%%%% TITLE
\title{nuScenes: A multimodal dataset for autonomous driving}

\author{
Holger Caesar, Varun Bankiti, Alex H. Lang, Sourabh Vora, Venice Erin Liong, Qiang Xu, \\
Anush Krishnan, Yu Pan, Giancarlo Baldan, Oscar Beijbom\\
nuTonomy: an APTIV company\\
{\tt\small nuscenes@nutonomy.com}
}

\maketitle
\thispagestyle{empty}

%%%%%%%%% ABSTRACT
\begin{abstract}
Robust detection and tracking of objects is crucial for the deployment of autonomous vehicle technology.
Image based benchmark datasets have driven development in computer vision tasks such as object detection, tracking and segmentation of agents in the environment.
Most autonomous vehicles, however, carry a combination of cameras and range sensors such as lidar and radar.
As machine learning based methods for detection and tracking become more prevalent,
there is a need to train and evaluate such methods on datasets containing range sensor data along with images.
In this work we present nuTonomy scenes (nuScenes), the first dataset to carry the full autonomous vehicle sensor suite: 6 cameras, 5 radars and 1 lidar, all with full 360 degree field of view.
nuScenes comprises 1000 scenes, each 20s long and fully annotated with 3D bounding boxes for 23 classes and 8 attributes.
It has 7x as many annotations and 100x as many images as the pioneering KITTI dataset.
We define novel 3D detection and tracking metrics.
We also provide careful dataset analysis as well as baselines for lidar and image based detection and tracking.
Data, development kit and more information are available online\footnote{\url{nuScenes.org}}.
\end{abstract}

%%%%%%%%% BODY TEXT
% !TEX root = ../nuscenes.tex

\vspace{-5mm}
\squeeze
\section{Introduction}
\label{sec:introduction}
\squeeze
\vspace{-2mm}

% Autonomous driving challenging problem
Autonomous driving has the potential to radically change the cityscape and save many human lives~\cite{changelives}.
A crucial part of safe navigation is the detection and tracking of agents in the environment surrounding the vehicle.
To achieve this, a modern self-driving vehicle deploys several sensors along with sophisticated detection and tracking algorithms.
Such algorithms rely increasingly on machine learning, which drives the need for benchmark datasets.
While there is a plethora of image datasets for this purpose (\tableref{tab:datasetcomparison}), there is a lack of multimodal datasets that exhibit the full set of challenges associated with building an autonomous driving perception system.
We released the nuScenes dataset to address this gap\footnote{\label{release_note}nuScenes teaser set released Sep. 2018, full release in March 2019.}.

\begin{figure}
\begin{center}
\vspace{-1mm}
\includegraphics[width=\linewidth]{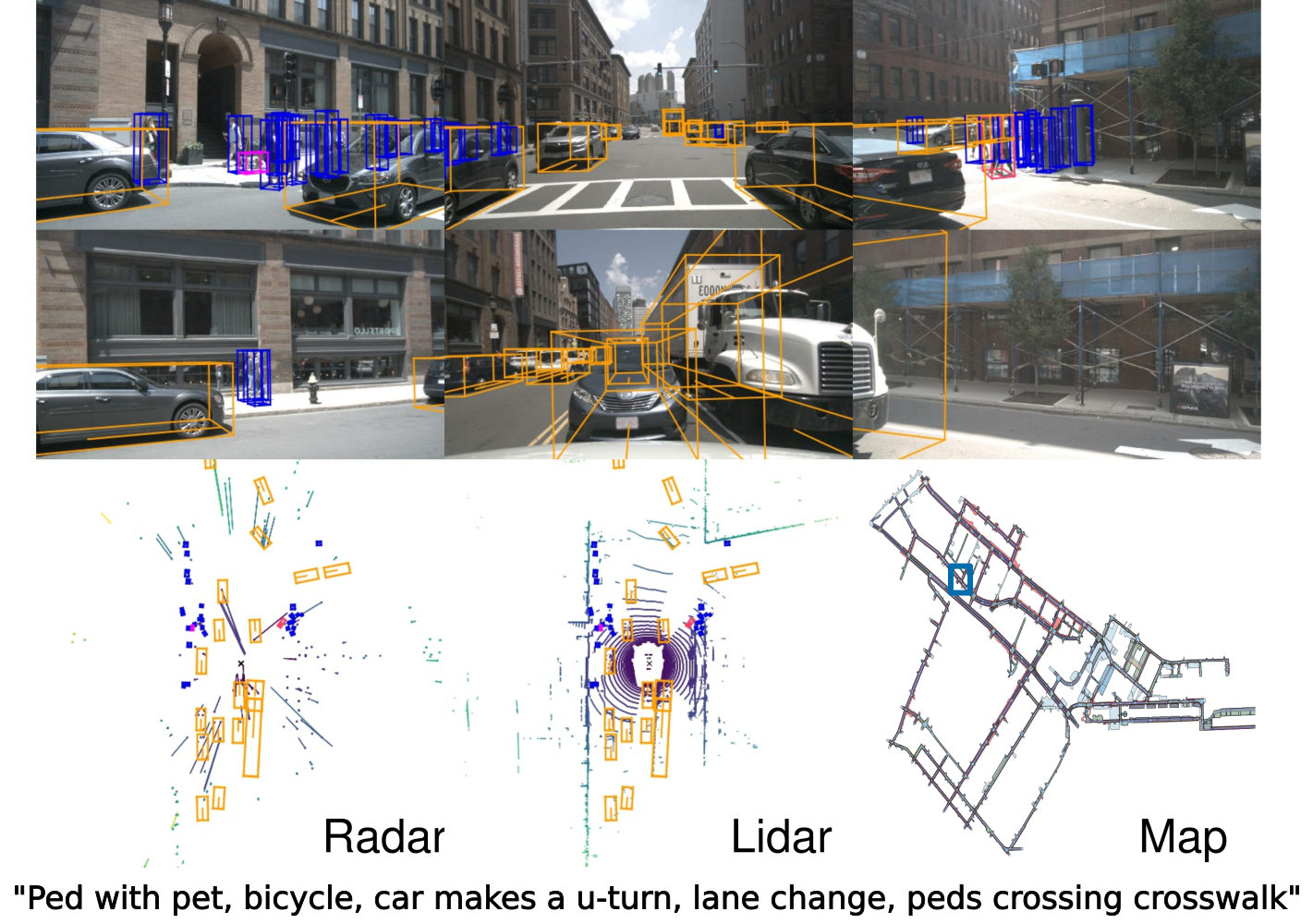}
\end{center}
\vspace{-3mm}
\caption{An example from the nuScenes dataset.
We see 6 different camera views, lidar and radar data, as well as the human annotated semantic map.
At the bottom we show the human written scene description.
}
\vspace{-4mm}
\label{fig:mainfigure}
\end{figure}

\begin{figure*}
\begin{center}
\includegraphics[width=\textwidth]{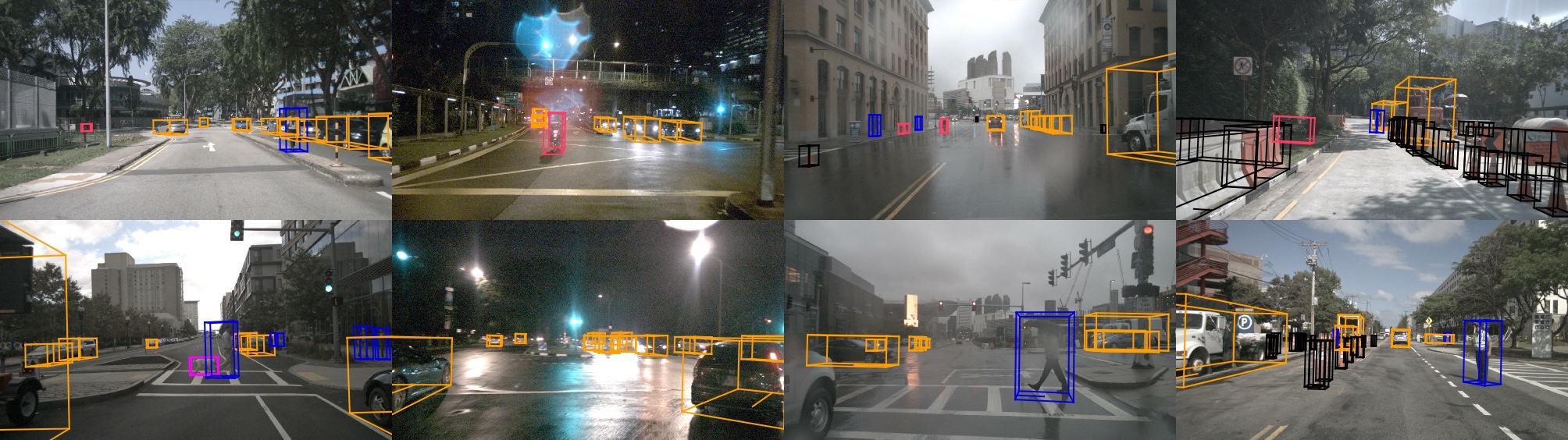}
\end{center}
\vspace{-4mm}
\caption{
Front camera images collected from clear weather (col 1), nighttime (col 2), rain (col 3) and construction zones (col 4).}
\label{fig:sample_images}
\end{figure*}

% Multimodal
Multimodal datasets are of particular importance as no single type of sensor is sufficient and the sensor types are complementary.
Cameras allow accurate measurements of edges, color and lighting enabling classification and localization on the image plane.
However, 3D localization from images is challenging~\cite{mono3d,3dop,deep3dbox,mlf_mono,oft,gpp,pseudo_lidar}.
Lidar pointclouds, on the other hand, contain less semantic information but highly accurate localization in 3D~\cite{pointpillars}.
Furthermore the reflectance of lidar is an important feature~\cite{struc_online_maps,pointpillars}.
However, lidar data is sparse and the range is typically limited to 50-150m.
Radar sensors achieve a range of 200-300m and measure the object velocity through the Doppler effect.
However, the returns are even sparser than lidar and less precise in terms of localization.
While radar has been used for decades~\cite{alessandretti2007vehicle,3decades_adas}, we are not aware of any autonomous driving datasets that provide radar data.

Since the three sensor types have different failure modes during difficult conditions, the joint treatment of sensor data is essential for agent detection and tracking.
Literature~\cite{sensorblocking} even suggests that multimodal sensor configurations are not just complementary, but provide redundancy in the face of sabotage, failure, adverse conditions and blind spots.
And while there are several works that have proposed fusion methods based on cameras and lidar~\cite{avod,mv3d,frustum,contfuse,pointfusion,sparsepool,multimodalsurvey}, PointPillars~\cite{pointpillars} showed a lidar-only method that performed on par with existing fusion based methods.
This suggests more work is required to combine multimodal measurements in a principled manner.

% Annotations
In order to train deep learning methods, quality data annotations are required.
Most datasets provide 2D semantic annotations as boxes or masks (class or instance)~\cite{camvid,cityscapes,mapillary,deepdrive,robotcar}.
At the time of the initial nuScenes release, only a few datasets annotated objects using 3D boxes~\cite{kitti,apolloscape, h3d}, and they did not provide the full sensor suite.
Following the nuScenes release, there are now several sets which contain the full sensor suite (\tableref{tab:datasetcomparison}).
Still, to the best of our knowledge, no other 3D dataset provides attribute annotations, such as pedestrian pose or vehicle state.

% Domains and nighttime
Existing AV datasets and vehicles are focused on particular operational design domains.
More research is required on generalizing to ``complex, cluttered and unseen environments''~\cite{driveability}.
Hence there is a need to study how detection methods generalize to different countries, lighting (daytime vs. nighttime), driving directions, road markings, vegetation, precipitation and previously unseen object types.

% Semantic maps
Contextual knowledge using semantic maps is also an important prior for scene understanding~\cite{hdnet, mapprior_tl, semantic_map_parking}.
For example, one would expect to find cars on the road, but not on the sidewalk or inside buildings.
With the notable exception of~\cite{lyftl5,argoverse}, most AV datasets do not provide semantic maps.

\begin{table*}
\setlength\tabcolsep{5.64pt} % Decrease this to reduce margins in each cell.
\center
\footnotesize
\begin{tabularx}{1\textwidth}{| L{2.1cm} | C{0.8cm} | C{0.67cm} | C{0.6cm} | C{0.8cm} | C{0.7cm} | C{0.7cm} | C{0.85cm} | C{0.78cm} | C{1.05cm} | C{0.8cm} | C{0.78cm} | C{1.5cm}|}
\hline
\textbf{Dataset} & \textbf{Year} & \textbf{Sce- nes} & \textbf{Size (hr)} & \textbf{RGB imgs} & \textbf{PCs lidar$^{\dagger\dagger}$} & \textbf{PCs radar} & \textbf{Ann. frames} & \textbf{3D boxes} & \textbf{Night / Rain} & \textbf{Map layers} & \textbf{Clas- ses}  & \textbf{Locations}\\
\hline
\hline
CamVid~\cite{camvid}								&	2008		&	4				&	0.4			&	18k			&	0			&	0			&	700			&	0			&	No/No			&	0     &   32   &   Cambridge\\
\hline
Cityscapes~\cite{cityscapes}						&	2016		&	n/a				&	-			&	25k			&	0			&	0			&	25k			&	0			&	No/No			&	0     &   30   &   50 cities\\
\hline
Vistas~\cite{mapillary}								&	2017		&	n/a				&	-			&	25k			&	0			&	0			&	25k			&	0			&	Yes/Yes         &	0     &  152   &   Global\\
\hline
BDD100K~\cite{deepdrive}							&	2017		&	100k 	        &	1k	        &	100M        &	0			&	0			&	100k		&	0			&	Yes/Yes         &	0     &   10   &   NY, SF\\
\hline
%M. obj. det.~\cite{multispectral_object} 			&	2017		&	-				&	-			&	7.5k		&	0			&	0			&	7.5k		&	0			&	Yes/No	        &	No     &    5   &   Tokyo\\
%\hline
%M. sem. seg.~\cite{multispectral_seg}				&	2017		&	-				&	-			&	1.6k		&	0			&	0			&	1.6k		&	0			&	Yes/No	        &	No     &    8   &   Tokyo\\
%\hline
ApolloScape~\cite{apolloscape}						&	2018		&	-				&	100			&	144k		&	0$^{**}$	&	0			&   144k        &   70k         &	Yes/No	        &	0    &   8-35 &   4x China\\
\hline
$D^2$-City~\cite{d2_city}                           &   2019        &   1k$^{\dag}$     &   -           &   700k$^{\dag}$&  0           &   0           &   700k$^{\dag}$&   0          &   No/Yes          &	0     &   12   &   5x China\\
\hline
\hline
KITTI~\cite{kitti}									&	2012		&	22				&	1.5			&	15k			&	15k			&	0			&	15k			&	200k		&	No/No			&	0     &    8   &   Karlsruhe\\
\hline
AS lidar~\cite{traffic_predict}						&	2018		&	-				&	2			&	0			&	20k			&	0			&	20k			& 	475k		&	-/-			    &	0     &    6   &   China\\
\hline
KAIST~\cite{kaist}									&	2018		&	-				&	-			&	8.9k		&	8.9k		&	0			&	8.9k		&	0			&	\textbf{Yes}/No	&	0     &    3   &   Seoul\\
\hline
H3D~\cite{h3d}										&	2019        &	160				&	0.77		&	83k		    &	27k			&	0 			&	27k			& 	1.1M		&	No/No			&	0     &    8   &   SF\\
\hline
nuScenes											&	2019		&	\textbf{1k}		&	5.5         &\textbf{1.4M}  &\textbf{400k}  &   \textbf{1.3M}&	\textbf{40k}& 1.4M &	\textbf{Yes/Yes}&\textbf{11}& \textbf{23}  &   Boston, SG\\
\hline
\hline
Argoverse~\cite{argoverse}                          &   2019       &   113$^{\dag}$     &   0.6$^{\dag}$ & 490k$^{\dag}$ &   44k        &   0           &   22k$^{\dag}$& 993k$^{\dag}$ & \textbf{Yes/Yes}  & 2      &   15    &   Miami, PT\\
\hline
Lyft L5~\cite{lyftl5}                               &   2019       &   366              &   2.5          &   323k        &   46k        &   0           & \textbf{46k}  & 1.3M &   No/No           &	7   &    9    &   Palo Alto\\
\hline
Waymo Open~\cite{waymo_open_dataset}                &   2019       &  \textbf{1k}       & 5.5   &  1M  & 200k  &   0           & \textbf{200k}$^{\ddagger}$ & \textbf{12M}$^{\ddagger}$ & \textbf{Yes/Yes}  &	0     &    4    &   3x USA\\
\hline
A$^*$3D~\cite{astar3d}                              &   2019       &   n/a              & \textbf{55}   &   39k         &   39k        &   0           & \textbf{39k}  &   230k        & \textbf{Yes/Yes}  &	0     &    7    &   SG\\
\hline
A2D2~\cite{audi_a2d2}                               &   2019       &   n/a              &   -            &   -           &   -          &   0           &   12k         &   -           &   -/-             & 	0     &   14    &   3x Germany\\
\hline
\end{tabularx}
\caption{
AV dataset comparison.
The top part of the table indicates datasets without range data.
The middle and lower parts indicate datasets (not publications) with range data released until and after the initial release of this dataset.
We use bold highlights to indicate the best entries in every column among the datasets with range data.
Only datasets which provide annotations for at least \emph{car}, \emph{pedestrian} and \emph{bicycle} are included in this comparison.
($^{\dag}$) We report numbers only for scenes annotated with cuboids.
($^{\ddagger}$) The current Waymo Open dataset size is comparable to nuScenes, but at a 5x higher annotation frequency.
($^{\dagger\dagger}$) Lidar pointcloud count collected from \emph{each lidar}.
(**)~\cite{apolloscape} provides static depth maps.
(-) indicates that no information is provided.
SG: Singapore, NY: New York, SF: San Francisco, PT: Pittsburgh, AS: ApolloScape.
}
\label{tab:datasetcomparison}
\end{table*}

% !TEX root = ../nuscenes.tex

\squeeze
\vspace{-1mm}
\subsection{Contributions}
\squeeze
\vspace{-2mm}
From the complexities of the multimodal 3D detection challenge, and the limitations of current AV datasets,
a large-scale multimodal dataset with \ang{360} coverage across all vision and range sensors collected from diverse situations alongside map information would boost AV scene-understanding research further.
nuScenes does just that, and it is the main contribution of this work.

nuScenes represents a large leap forward in terms of data volumes and complexities (\tableref{tab:datasetcomparison}),
and is the first dataset to provide \ang{360} sensor coverage from the \emph{entire sensor suite}.
It is also the first AV dataset to include \emph{radar data} and captured using an AV \emph{approved for public roads}.
It is further the first multimodal dataset that contains data from \emph{nighttime} and \emph{rainy} conditions,
and with \emph{object attributes and scene descriptions} in addition to object class and location.
Similar to~\cite{woodscape}, nuScenes is a holistic scene understanding benchmark for AVs.
It enables research on multiple tasks such as object detection, tracking and behavior modeling in a range of conditions.

Our second contribution is new detection and tracking metrics aimed at the AV application.
We train 3D object detectors and trackers as a baseline, including a novel approach of using multiple lidar sweeps to enhance object detection.
We also present and analyze the results of the nuScenes object detection and tracking challenges.

Third, we publish the devkit, evaluation code, taxonomy, annotator instructions, and database schema for industry-wide standardization.
Recently, the Lyft L5~\cite{lyftl5} dataset adopted this format to achieve compatibility between the different datasets.
The nuScenes data is published under CC BY-NC-SA 4.0 license, which means that anyone can use this dataset for non-commercial research purposes.
All data, code, and information is made available online\footnote{\url{github.com/nutonomy/nuscenes-devkit}}.

Since the release, nuScenes has received strong interest from the AV community~\cite{megvii, monodis, autonue, objspecificdistance, spagnn-uber, monoloco, precog, datasetsurvey1, datasetsurvey2, datasetsurvey3, datasetsurvey4}.
Some works extended our dataset to introduce new annotations for natural language object referral~\cite{talk2car} and high-level scene understanding~\cite{scene-understanding}.
The detection challenge enabled lidar based and camera based detection works such as~\cite{megvii, monodis},
that improved over the state-of-the-art at the time of initial release~\cite{pointpillars,oft} by $40\%$ and $81\%$ (\tableref{table:detection_challenge}).
nuScenes has been used for 3D object detection~\cite{sarpnet, howmuchrealdata}, multi-agent forecasting~\cite{spagnn-uber, precog}, pedestrian localization~\cite{monoloco}, weather augmentation~\cite{rain}, and  moving pointcloud prediction~\cite{pointrnn}.
Being still the only annotated AV dataset to provide radar data, nuScenes encourages researchers to explore radar and sensor fusion for object detection~\cite{pointrnn, rvnet, pointpainting}.

\vspace{-1mm}
\squeeze
\subsection{Related datasets}
\label{sec:relatedwork}
\squeeze
\vspace{-2mm}

% Image based datasets
The last decade has seen the release of several driving datasets which have played a huge role in scene-understanding research for AVs.
Most datasets have focused on 2D annotations (boxes, masks) for RGB camera images.
CamVid~\cite{camvid}, Cityscapes~\cite{cityscapes}, Mapillary Vistas~\cite{mapillary}, $D^2$-City~\cite{d2_city}, BDD100k~\cite{deepdrive} and Apolloscape~\cite{apolloscape} released ever growing datasets with segmentation masks.
Vistas, $D^2$-City and BDD100k also contain images captured during different weather and illumination settings.
Other datasets focus exclusively on pedestrian annotations on images~\cite{hog,eth_persons,tud_brussels_ped,daimler_ped,citypersons,caltech_ped,nightowl}.
The ease of capturing and annotating RGB images have made the release of these large image-only datasets possible.

% Multimodal datasets
On the other hand, multimodal datasets, which are typically comprised of images, range sensor data (lidars, radars), and GPS/IMU data,
are expensive to collect and annotate due to the difficulties of integrating, synchronizing, and calibrating multiple sensors. KITTI~\cite{kitti} was the pioneering multimodal dataset providing dense pointclouds from a lidar sensor as well as front-facing stereo images and GPS/IMU data.
It provides 200k 3D boxes over 22 scenes which helped advance the state-of-the-art in 3D object detection.
The recent H3D dataset~\cite{h3d} includes 160 crowded scenes with a total of 1.1M 3D boxes annotated over 27k frames.
The objects are annotated in the full \ang{360} view, as opposed to KITTI where an object is only annotated if it is present in the frontal view.
The KAIST multispectral dataset~\cite{kaist} is a multimodal dataset that consists of RGB and thermal camera, RGB stereo, 3D lidar and GPS/IMU.
It provides nighttime data, but the size of the dataset is limited and annotations are in 2D.
Other notable multimodal datasets include~\cite{lidar_video} providing driving behavior labels, \cite{3d_outdoor} providing place categorization labels and~\cite{malaga_urban, robotcar} providing raw data without semantic labels.

% Synthetic datasets
% An alternative to collecting real-world multimodal driving data is by generating synthetic data via simulation~\cite{carla, synthia, virtualkitti, playingforbenchmarks}.
% These have the advantage of simulating arbitrarily situations whilst avoiding the cost of human annotation.
% While these have made impressive progress, it is still difficult to photo-realistically simulate all the rare classes and scenarios that an AV may encounter in the real-world.

% New datasets
After the initial nuScenes release, \cite{waymo_open_dataset, argoverse, astar3d, audi_a2d2, lyftl5} followed to release their own large-scale AV datasets (\tableref{tab:datasetcomparison}).
Among these datasets, only the Waymo Open dataset~\cite{waymo_open_dataset} provides significantly more annotations,
mostly due to the higher annotation frequency ($10\text{Hz}$ vs. $2\text{Hz}$)\footnote{In preliminary analysis we found that annotations at $2\text{Hz}$ are robust to interpolation to finer temporal resolution, like $10\text{Hz}$ or $20\text{Hz}$.
A similar conclusion was drawn for H3D~\cite{h3d} where annotations are interpolated from $2\text{Hz}$ to $10\text{Hz}$.}.
A*3D takes an orthogonal approach where a similar number of frames (39k) are selected and annotated from 55 hours of data.
The Lyft L5 dataset~\cite{lyftl5} is most similar to nuScenes.
It was released using the nuScenes database schema and can therefore be parsed using the nuScenes devkit.

% !TEX root = ../nuscenes.tex

\vspace{-2mm}
\squeeze
\section{The nuScenes dataset}
\label{sec:dataset}
\squeeze
\vspace{-2mm}
Here we describe how we plan drives, setup our vehicles, select interesting scenes, annotate the dataset and protect the privacy of third parties.

\mypar{Drive planning.}
We drive in Boston (Seaport and South Boston) and Singapore (One North, Holland Village and Queenstown),
two cities that are known for their dense traffic and highly challenging driving situations.
We emphasize the diversity across locations in terms of vegetation, buildings, vehicles, road markings and right versus left-hand traffic.
From a large body of training data we manually select 84 logs with 15h of driving data (242km travelled at an average of 16km/h).
Driving routes are carefully chosen to capture a diverse set of locations (urban, residential, nature and industrial), times (day and night) and weather conditions (sun, rain and clouds).

\begin{figure}[b!]
\vspace{-3mm}
\begin{center}
\includegraphics[width=\linewidth]{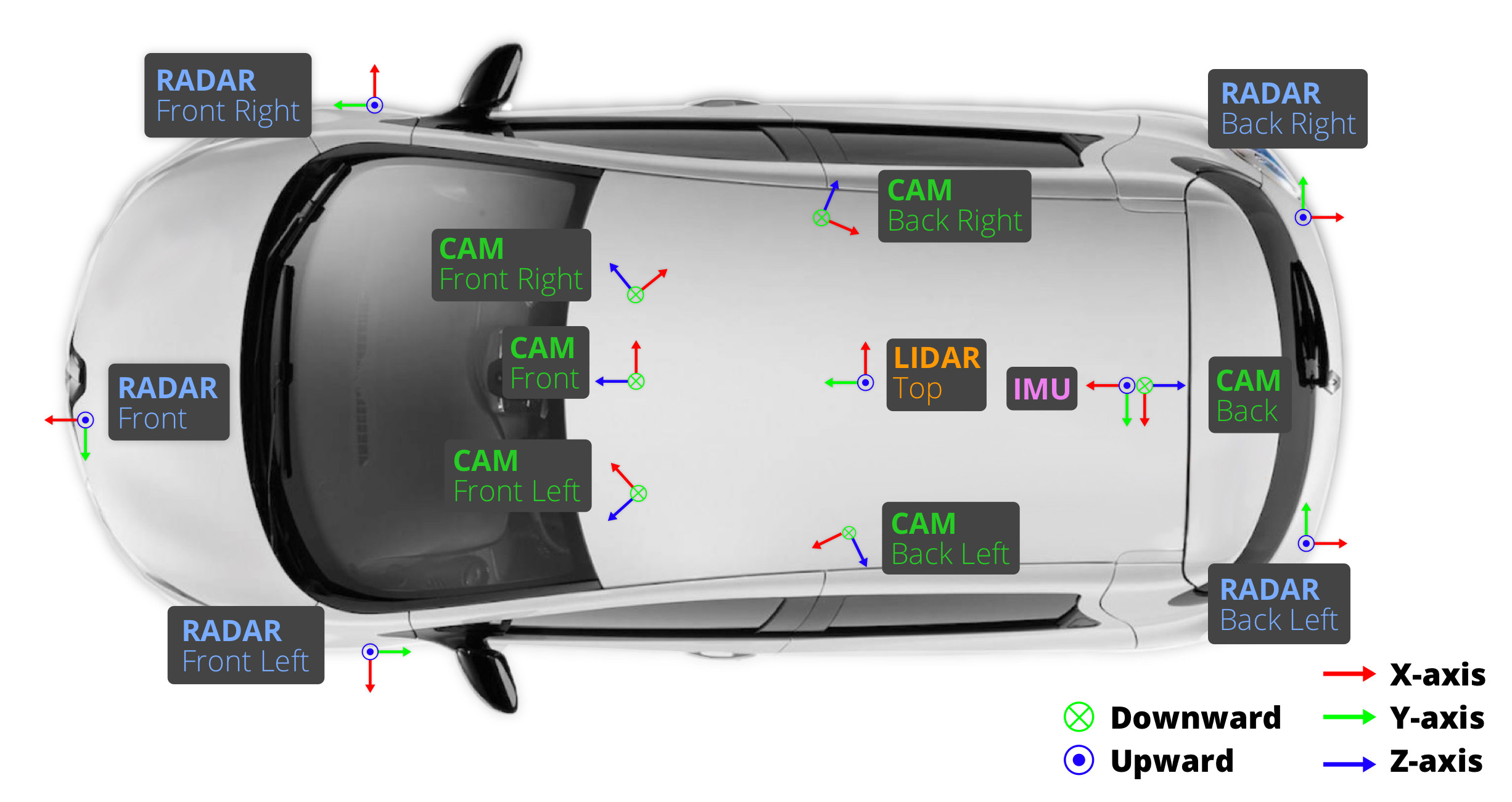}
\end{center}
\squeeze
\vspace{-4mm}
\caption{Sensor setup for our data collection platform.
}
\vspace{-2mm}
\label{fig:sensors}
\end{figure}

\mypar{Car setup.}
We use two Renault Zoe supermini electric cars with an identical sensor layout to drive in Boston and Singapore.
See \figref{fig:sensors} for sensor placements and \tableref{tab:sensors} for sensor details.
\begin{table}
\footnotesize
\centering
\begin{tabularx}{\linewidth}{p{1.5cm} | p{6.2cm}}
\textbf{Sensor} &   \textbf{Details}\\
\hline
6x Camera       &   RGB, $12\text{Hz}$ capture frequency, $1/1.8"$ CMOS sensor, $1600\times900$ resolution, auto exposure, JPEG compressed\\
1x Lidar        &   Spinning, $32$ beams, $20\text{Hz}$ capture frequency, \ang{360} horizontal FOV, \ang{-30} to \ang{+10} vertical FOV, $\leq 70m$ range, $\pm 2\text{cm}$ accuracy, up to $1.4M$ points per second.\\
5x Radar        &   $\leq 250m$ range, $77\text{GHz}$, FMCW, $13\text{Hz}$ capture frequency, $\pm 0.1\text{km/h}$ vel. accuracy\\
GPS \& IMU      &   GPS, IMU, AHRS. \ang{0.2} heading, \ang{0.1} roll/pitch, 20mm RTK positioning, 1000Hz update rate
\end{tabularx}
\vspace{-1mm}
\caption{
Sensor data in nuScenes.
}
\vspace{-2mm}
\label{tab:sensors}
\end{table}
Front and side cameras have a \ang{70} FOV and are offset by \ang{55}.
The rear camera has a FOV of \ang{110}.

\mypar{Sensor synchronization.}
To achieve good cross-modality data alignment between the lidar and the cameras,
the exposure of a camera is triggered when the top lidar sweeps across the center of the camera's FOV.
The timestamp of the image is the exposure trigger time; and the timestamp of the lidar scan is the time when the full rotation of the current lidar frame is achieved.
Given that the camera's exposure time is nearly instantaneous, this method generally yields good data alignment\footnote{The cameras run at $12\text{Hz}$ while the lidar runs at $20\text{Hz}$.
The $12$ camera exposures are spread as evenly as possible across the $20$ lidar scans, so not all lidar scans have a corresponding camera frame.}.
We perform motion compensation using the localization algorithm described below.

\mypar{Localization.}
Most existing datasets provide the vehicle location based on GPS and IMU~\cite{kitti,apolloscape,cityscapes,h3d}.
Such localization systems are vulnerable to GPS outages, as seen on the KITTI dataset~\cite{kitti,kittigps}.
As we operate in dense urban areas, this problem is even more pronounced.
To accurately localize our vehicle, we create a detailed HD map of lidar points in an offline step.
While collecting data, we use a Monte Carlo Localization scheme from lidar and odometry information~\cite{localization}.
This method is very robust and we achieve localization errors of $\leq 10cm$.
To encourage robotics research, we also provide the raw CAN bus data (e.g. velocities, accelerations, torque, steering angles, wheel speeds) similar to~\cite{hondadataset}.

\begin{figure}
\begin{center}
\includegraphics[width=\linewidth]{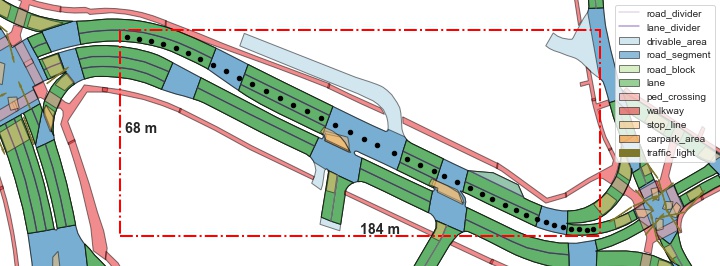}
\end{center}
\squeeze
\vspace{-10pt}
\caption{Semantic map of nuScenes with 11 semantic layers in different colors.
To show the path of the ego vehicle we plot each keyframe ego pose from \emph{scene-0121} with black spheres.
}
\vspace{-1mm}
\label{fig:semantic-map}
\end{figure}

\mypar{Maps.}
We provide highly accurate human-annotated semantic maps of the relevant areas.
The original rasterized map includes only roads and sidewalks with a resolution of $10\text{px/m}$.
The vectorized \emph{map expansion} provides information on 11 semantic classes as shown in \figref{fig:semantic-map}, making it richer than the semantic maps of other datasets published since the original release~\cite{argoverse,lyftl5}.
% Specifically, Argoverse contains 2 map layers: lane centerlines with attributes (i.e. if in intersection, turn direction) and driveable area or region of interest.
% They also have another map for ground height from 1.0m resolution.
%On the other hand, Lyft provides a semantic map consisting of road segment lanes, junction lanes, pedestrian crosswalks, stop signs, parking zones, speed bumps, speed humps.
We encourage the use of localization and semantic maps as strong priors for all tasks.
%(e.g. pedestrians are typically found on sidewalks or crosswalks)
Finally, we provide the baseline routes - the idealized path an AV \emph{should} take, assuming there are no obstacles.
This route may assist trajectory prediction~\cite{precog}, as it simplifies the problem by reducing the search space of viable routes.

\mypar{Scene selection.}
After collecting the raw sensor data, we manually select $1000$ \emph{interesting} scenes of $20s$ duration each.
Such scenes include high traffic density (e.g. intersections, construction sites), rare classes (e.g. ambulances, animals), potentially dangerous traffic situations (e.g. jaywalkers, incorrect behavior), maneuvers (e.g. lane change, turning, stopping) and situations that may be difficult for an AV.
We also select some scenes to encourage diversity in terms of spatial coverage, different scene types, as well as different weather and lighting conditions.
Expert annotators write textual descriptions or \emph{captions} for each scene
(e.g.: ``Wait at intersection, peds on sidewalk, bicycle crossing, jaywalker, turn right, parked cars, rain'').
%Each caption is a list of keywords in chronological order, describing why the scene is interesting,
%We provide our annotators a list of frequently occurring keywords,
%but encourage the definition of new keywords for previously unseen situations, e.g. ``lightning'', ``tricycle'' or ``barrier being pushed across crosswalk''.
%We additionally include the keywords ``rain'' and ``night'' to enable users to analyze these more difficult scene types separately.

\mypar{Data annotation.}
Having selected the scenes, we sample keyframes (image, lidar, radar) at $2\text{Hz}$.
We annotate each of the 23 object classes in every keyframe with a semantic category, attributes (visibility, activity, and pose) and a cuboid modeled as x, y, z, width, length, height and yaw angle.
We annotate objects continuously throughout each scene if they are covered by at least one lidar or radar point.
Using expert annotators and multiple validation steps, we achieve highly accurate annotations.
We also release intermediate sensor frames, which are important for tracking, prediction and object detection as shown in \secref{sec:experiments:analysis}.
At capture frequencies of $12\text{Hz}$, $13\text{Hz}$ and $20\text{Hz}$ for camera, radar and lidar, this makes our dataset unique.
Only the Waymo Open dataset provides a similarly high capture frequency of $10\text{Hz}$.

\mypar{Annotation statistics.}
Our dataset has 23 categories including different vehicles, types of pedestrians, mobility devices and other objects (\figref{fig:annotation_count}-SM).
We present statistics on geometry and frequencies of different classes (\figref{fig:basic_stats}-SM).
Per keyframe there are 7 pedestrians and 20 vehicles on average.
Moreover, 40k keyframes were taken from four different scene locations (Boston: 55\%, SG-OneNorth: 21.5\%, SG-Queenstown: 13.5\%, SG-HollandVillage: 10\%) with various weather and lighting conditions (rain: 19.4\%, night: 11.6\%).
Due to the finegrained classes in nuScenes, the dataset shows severe class imbalance with a ratio of 1:10k for the least and most common class annotations (1:36 in KITTI).
%Even in our detection challenge, the \emph{bicycle} class have ~17 and ~40 times less annotations than pedestrians and vehicles, as compared to KITTI which have only ~14 and ~7 times less.
This encourages the community to explore this long tail problem in more depth.

\begin{figure}
\begin{center}
\vspace{-1mm}
\includegraphics[width=\linewidth]{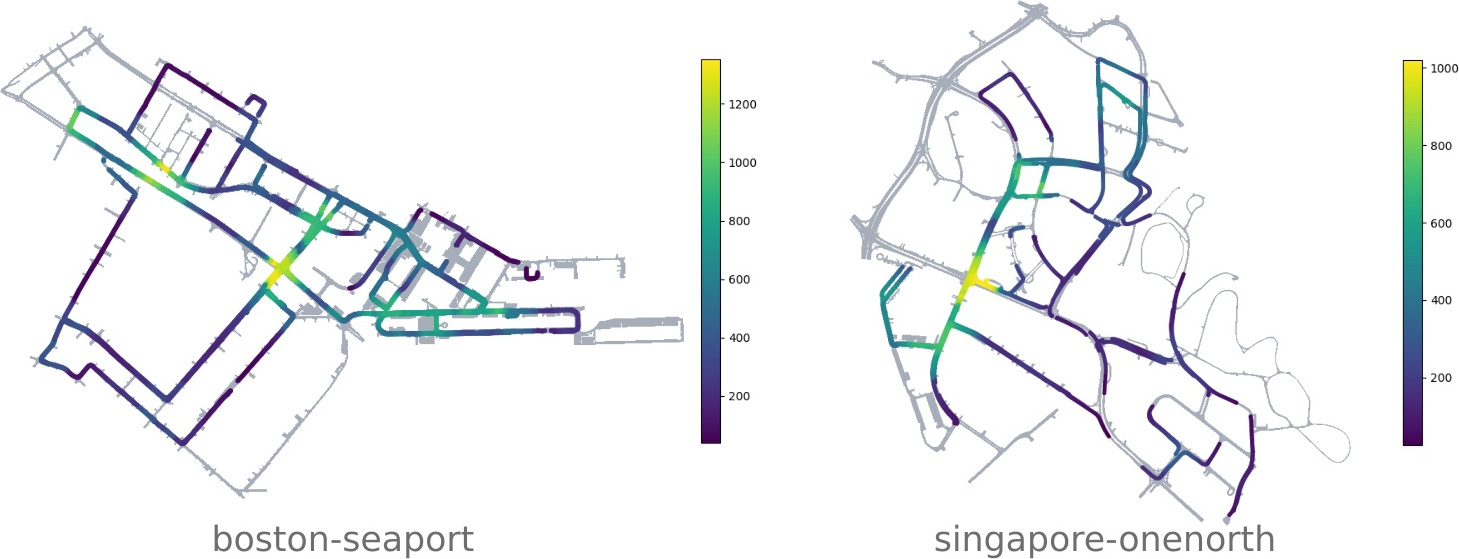}
\end{center}
\squeeze
\vspace{-1mm}
\caption{
Spatial data coverage for two nuScenes locations.
Colors indicate the number of keyframes with ego vehicle poses within a 100m radius across all scenes.}
\vspace{-1mm}
\label{fig:analysiscoverage}
\end{figure}

\figref{fig:analysiscoverage} shows spatial coverage across all scenes.
We see that most data comes from intersections.
\figref{fig:dist_orient_hists}-SM shows that \emph{car} annotations are seen at varying distances and as far as 80m from the ego-vehicle.
Box orientation is also varying, with the most number in vertical and horizontal angles for cars as expected due to parked cars and cars in the same lane.
Lidar and radar points statistics inside each box annotation are shown in \figref{fig:lidar_pt_hists}-SM.
Annotated objects contain up to 100 lidar points even at a radial distance of 80m and at most 12k lidar points at 3m.
At the same time they contain up to 40 radar returns at 10m and 10 at 50m.
The radar range far exceeds the lidar range at up to 200m.

% !TEX root = ../nuscenes.tex

\vspace{-0mm}
\squeeze
\section{Tasks \& Metrics}
\label{sec:tasks}
\squeeze
\vspace{-1mm}
The multimodal nature of nuScenes supports a multitude of tasks including detection, tracking, prediction \& localization.
Here we present the detection and tracking tasks and metrics.
We define the \emph{detection} task to only operate on sensor data between $[t-0.5, t]$ seconds for an object at time $t$,
whereas the \emph{tracking} task operates on data between $[0, t]$.

\vspace{-1mm}
\squeeze
\subsection{Detection}
\label{sec:tasks:detection}
\squeeze
\vspace{-1mm}
The nuScenes detection task requires detecting 10 object classes with 3D bounding boxes, attributes (e.g. sitting vs. standing), and velocities.
The 10 classes are a subset of all 23 classes annotated in nuScenes (\tableref{table:class_map}-SM).

\mypar{Average Precision metric.}
We use the Average Precision (AP) metric~\cite{kitti, pascal}, but define a match by thresholding the 2D center distance $d$ on the ground plane instead of intersection over union (IOU).
This is done in order to decouple detection from object size and orientation but also because objects with small footprints, like pedestrians and bikes,
if detected with a small translation error, give $0$ IOU (\figref{fig:matching_study}).
This makes it hard to compare the performance of vision-only methods which tend to have large localization errors~\cite{oft}.

We then calculate AP as the normalized area under the precision recall curve for recall and precision over $10\%$.
Operating points where recall or precision is less than 10\% are removed in order to minimize the impact of noise commonly seen in low precision and recall regions.
If no operating point in this region is achieved, the AP for that class is set to zero.
We then average over matching thresholds of $\mathbb{D} = \{0.5, 1, 2, 4\}$ meters and the set of classes $\mathbb{C}$:
\begin{align}
\vspace{+1mm}
\textrm{mAP} = \frac{1}{|\mathbb{C}| |\mathbb{D}|} \sum_{c \in \mathbb{C}} \sum_{d \in \mathbb{D}} \textrm{AP}_{c, d} \label{eq:map}
\vspace{-4mm}
\end{align}

\mypar{True Positive metrics.}
In addition to AP, we measure a set of \emph{True Positive metrics} (TP metrics) for each prediction that was matched with a ground truth box.
All TP metrics are calculated using $d=2$m center distance during matching, and they are all designed to be positive scalars.
In the proposed metric, the TP metrics are all in native units (see below) which makes the results easy to interpret and compare.
Matching and scoring happen independently per class and each metric is the average of the cumulative mean at each achieved recall level above $10\%$.
If $10\%$ recall is not achieved for a particular class, all TP errors for that class are set to $1$.
The following TP errors are defined:

Average Translation Error (ATE) is the Euclidean center distance in 2D (units in $meters$).
Average Scale Error (ASE) is the 3D intersection over union (IOU) after aligning orientation and translation ($1-IOU$).
Average Orientation Error (AOE) is the smallest yaw angle difference between prediction and ground truth ($radians$).
All angles are measured on a full $360^{\circ}$ period except for barriers where they are measured on a $180^{\circ}$ period.
Average Velocity Error (AVE) is the absolute velocity error as the L2 norm of the velocity differences in 2D ($m/s$).
Average Attribute Error (AAE) is defined as 1 minus attribute classification accuracy ($1-acc$).
For each TP metric we compute the mean TP metric ($\textrm{mTP}$) over all classes:
\begin{align}
\textrm{mTP} = \frac{1}{|\mathbb{C}|} \sum_{c \in \mathbb{C} } \textrm{TP}_{c}
\label{eq:mtp}
\end{align}
\vspace{-3mm}

We omit measurements for classes where they are not well defined:
AVE for cones and barriers since they are stationary;
AOE of cones since they do not have a well defined orientation;
and AAE for cones and barriers since there are no attributes defined on these classes.

\mypar{nuScenes detection score.}
mAP with a threshold on IOU is perhaps the most popular metric for object detection~\cite{kitti,cityscapes,imagenet}.
However, this metric can not capture all aspects of the nuScenes detection tasks, like velocity and attribute estimation.
Further, it couples location, size and orientation estimates.
The ApolloScape~\cite{apolloscape} 3D car instance challenge disentangles these by defining thresholds for each error type and recall threshold.
This results in $10 \times 3$ thresholds, making this approach complex, arbitrary and unintuitive.
% Recently, Waymo~\cite{waymo} suggested an mAP weighted by heading metric.
We propose instead consolidating the different error types into a scalar score: the nuScenes detection score (NDS).
\begin{align}
\textrm{NDS} = \frac{1}{10} [5~\textrm{mAP} + \sum_{\textrm{mTP} \in \mathbb{TP}} ( 1 - \min(1, \; \textrm{mTP}) ) ] \label{eq:nds}
\end{align}
Here mAP is mean Average Precision~\eqref{eq:map},  and $\mathbb{TP}$ the set of the five mean True Positive metrics~\eqref{eq:mtp}.
Half of NDS is thus based on the detection performance while the other half quantifies the quality of the detections in terms of box location, size, orientation, attributes, and velocity.
Since mAVE, mAOE and mATE can be larger than $1$, we bound each metric between $0$ and $1$ in \eqref{eq:nds}.

\squeeze
\vspace{+1mm}
\subsection{Tracking}
\label{sec:tasks:tracking-task}
\squeeze
In this section we present the tracking task setup and metrics.
The focus of the tracking task is to track all detected objects in a scene.
All detection classes defined in \secref{sec:tasks:detection} are used, except the static classes: \emph{barrier}, \emph{construction} and \emph{trafficcone}.

\vspace{+1mm}
\mypar{AMOTA and AMOTP metrics.}
Weng and Kitani~\cite{ab3dmot} presented a similar 3D MOT benchmark on KITTI~\cite{kitti}.
They point out that traditional metrics do not take into account the confidence of a prediction.
Thus they develop Average Multi Object Tracking Accuracy (AMOTA) and Average Multi Object Tracking Precision (AMOTP), which average MOTA and MOTP across all recall thresholds.
By comparing the KITTI and nuScenes leaderboards for detection and tracking, we find that nuScenes is significantly more difficult.
Due to the difficulty of nuScenes, the traditional MOTA metric is often zero.
In the updated formulation $\text{sMOTA}_r$\cite{ab3dmot}\footnote{Pre-prints of this work referred to $\text{sMOTA}_r$ as MOTAR.},
MOTA is therefore augmented by a term to adjust for the respective recall:

{
\vspace{-2mm}
\begin{footnotesize}
\begin{align*} % align* hides the equation numbering
  \mathit{sMOTA}_r &= \max{\left( 0,\; 1 \, - \, \frac{\mathit{IDS}_r + \mathit{FP}_r + \mathit{FN}_r - (1-r) \mathit{P}}{r \mathit{P}} \right)}
\end{align*}
\end{footnotesize}
\vspace{+1mm}
}
This is to guarantee that $\text{sMOTA}_r$ values span the entire $[0, 1]$ range.
We perform 40-point interpolation in the recall range $[0.1, 1]$ (the recall values are denoted as $\mathcal{R}$).
The resulting sAMOTA metric is the main metric for the tracking task:
{
\begin{footnotesize}
\begin{align*} % align* hides the equation numbering
  \mathit{sAMOTA} &= \frac{1}{\vert\mathcal{R}\vert} \, \sum_{r \in \mathcal{R}} \mathit{sMOTA}_r \\
\end{align*}
\end{footnotesize}
}

\vspace{-6mm}
\mypar{Traditional metrics.}
We also use traditional tracking metrics such as MOTA and MOTP~\cite{motmetrics}, false alarms per frame, mostly tracked trajectories, mostly lost trajectories, false positives, false negatives, identity switches, and track fragmentations.
Similar to~\cite{ab3dmot}, we try all recall thresholds and then use the threshold that achieves highest $\text{sMOTA}_r$.

\mypar{TID and LGD metrics.}
In addition, we devise two novel metrics: Track initialization duration (TID) and longest gap duration (LGD).
Some trackers require a fixed window of past sensor readings or perform poorly without a good initialization.
TID measures the duration from the beginning of the track until the time an object is first detected.
LGD computes the longest duration of \emph{any} detection gap in a track.
If an object is not tracked, we assign the entire track duration as TID and LGD.
For both metrics, we compute the average over all tracks.
These metrics are relevant for AVs as many short-term track fragmentations may be more acceptable than missing an object for several seconds.

% !TEX root = ../nuscenes.tex

\vspace{+2mm}
\squeeze
\section{Experiments}
\label{sec:experiments}
\squeeze
\vspace{-1mm}
In this section we present object detection and tracking experiments on the nuScenes dataset, analyze their characteristics and suggest avenues for future research.

\squeeze
\subsection{Baselines}
\label{sec:experiments:baselines}
\squeeze
We present a number of baselines with different modalities for detection and tracking.

\vspace{+2mm}
\mypar{Lidar detection baseline.}
\label{sec:experiments:lidar-baseline}

To demonstrate the performance of a leading algorithm on nuScenes, we train a lidar-only 3D object detector, PointPillars~\cite{pointpillars}.
We take advantage of temporal data available in nuScenes by accumulating lidar sweeps for a richer pointcloud as input.
A single network was trained for all classes.
The network was modified to also learn velocities as an additional regression target for each 3D box.
We set the box attributes to the most common attribute for each class in the training data.

\mypar{Image detection baseline.}
To examine image-only 3D object detection, we re-implement the Orthographic Feature Transform (OFT)~\cite{oft} method.
A single OFT network was used for all classes.
We modified the original OFT to use a SSD detection head and confirmed that this matched published results on KITTI.
The network takes in a single image from which the full \ang{360} predictions are combined together from all 6 cameras using non-maximum suppression (NMS).
We set the box velocity to zero and attributes to the most common attribute for each class in the train data.

\mypar{Detection challenge results.}
We compare the results of the top submissions to the nuScenes detection challenge 2019.
Among all submissions, Megvii~\cite{megvii} gave the best performance.
It is a lidar based class-balanced multi-head network with sparse 3D convolutions.
Among image-only submissions, MonoDIS~\cite{monodis} was the best, significantly outperforming our image baseline and even some lidar based methods.
It uses a novel disentangling 2D and 3D detection loss.
Note that the top methods all performed importance sampling, which shows the importance of addressing the class imbalance problem.

\mypar{Tracking baselines.}
We present several baselines for tracking from camera and lidar data.
From the detection challenge, we pick the best performing lidar method (Megvii~\cite{megvii}), the fastest reported method at inference time (PointPillars~\cite{pointpillars}), as well as the best performing camera method (MonoDIS~\cite{monodis}).
Using the detections from each method, we setup baselines using the tracking approach described in~\cite{ab3dmot}.
We provide detection and tracking results for each of these methods on the train, val and test splits to facilitate more systematic research.
See the Supplementary Material for the results of the 2019 nuScenes tracking challenge.

\squeeze
\vspace{+3mm}
\subsection{Analysis}
\label{sec:experiments:analysis}
\squeeze
Here we analyze the properties of the methods presented in \secref{sec:experiments:baselines}, as well as the dataset and matching function.

\begin{figure}
\begin{center}
%\vspace{-2mm}
\includegraphics[width=0.9\linewidth]{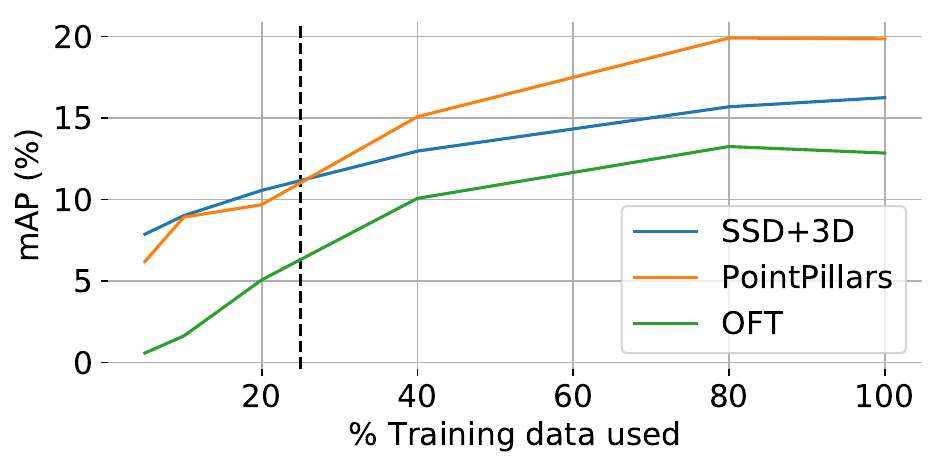}
\end{center}
\squeeze
\vspace{-3mm}
\caption{Amount of training data vs. mean Average Precision (mAP) on the val set of nuScenes.
The dashed black line corresponds to the amount of training data in KITTI~\cite{kitti}.
}
\vspace{+1mm}
\label{fig:train_size_study}
\end{figure}

\mypar{The case for a large benchmark dataset.}
One of the contributions of nuScenes is the dataset size, and in particular the increase compared to KITTI (\tableref{tab:datasetcomparison}).
Here we examine the benefits of the larger dataset size.
We train PointPillars~\cite{pointpillars}, OFT~\cite{oft} and an additional image baseline, SSD+3D, with varying amounts of training data.
SSD+3D has the same 3D parametrization as MonoDIS~\cite{monodis}, but use a single stage design~\cite{ssd}.
For this ablation study we train PointPillars with 6x fewer epochs and a one cycle optimizer schedule~\cite{onecycle} to cut down the training time.
Our main finding is that the \emph{method ordering changes} with the amount of data (\figref{fig:train_size_study}).
In particular, PointPillars performs similar to SSD+3D at data volumes commensurate with KITTI, but as more data is used, it is clear that PointPillars is stronger.
This suggests that the full potential of complex algorithms can only be verified with a bigger and more diverse training set.
A similar conclusion was reached by~\cite{lasernet, starnet} with~\cite{starnet} suggesting that the KITTI leaderboard reflects the data aug. method rather than the actual algorithms.

\begin{figure}
\begin{center}
\includegraphics[width=0.9\linewidth]{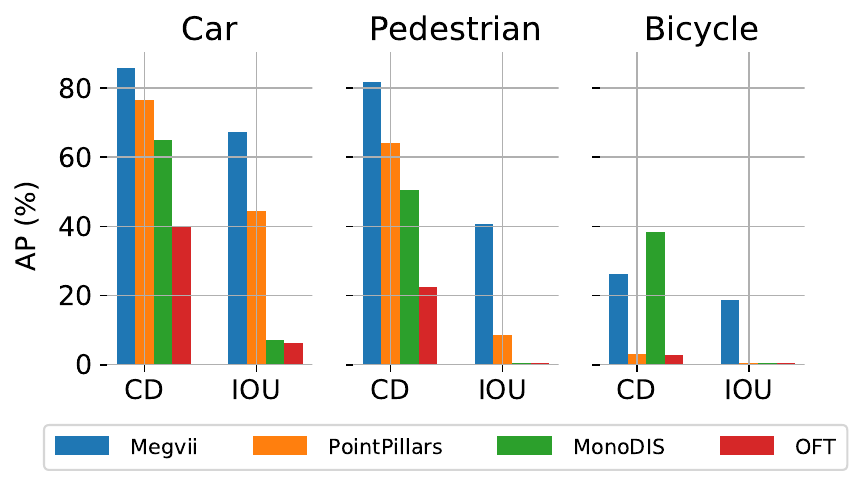}
\end{center}
\squeeze
\vspace{-3mm}
\caption{
Average precision vs. matching function.
CD: Center distance.
IOU: Intersection over union.
We use $\text{IOU} = 0.7$ for \emph{car} and $\text{IOU} = 0.5$ for \emph{pedestrian} and \emph{bicycle} following KITTI~\cite{kitti}.
We use $\text{CD} = 2m$ for the TP metrics in \secref{sec:tasks:detection}.
}
\label{fig:matching_study}
\end{figure}

\mypar{The importance of the matching function.}
We compare performance of published methods (\tableref{table:detection_challenge}) when using our proposed 2m center-distance matching versus the IOU matching used in KITTI.
As expected, when using IOU matching, small objects like pedestrians and bicycles fail to achieve above 0 AP, making ordering impossible (\figref{fig:matching_study}).
In contrast, center distance matching declares MonoDIS a clear winner.
The impact is smaller for the car class, but also in this case it is hard to resolve the difference between MonoDIS and OFT.

The matching function also changes the balance between lidar and image based methods.
In fact, the ordering switches when using center distance matching to favour MonoDIS over both lidar based methods on the bicycle class (\figref{fig:matching_study}).
This makes sense since the thin structures of bicycles make them difficult to detect in lidar.
We conclude that center distance matching is more appropriate to rank image based methods alongside lidar based methods.

\mypar{Multiple lidar sweeps improve performance.}
According to our evaluation protocol (\secref{sec:tasks:detection}), one is only allowed to use $0.5s$ of previous data to make a detection decision.
This corresponds to 10 previous lidar sweeps since the lidar is sampled at $20\text{Hz}$.
We device a simple way of incorporating multiple pointclouds into the PointPillars baseline and investigate the performance impact.
Accumulation is implemented by moving all pointclouds to the coordinate system of the keyframe and appending a scalar time-stamp to each point indicating the time delta in seconds from the keyframe.
The encoder includes the time delta as an extra decoration for the lidar points.
Aside from the advantage of richer pointclouds, this also provides temporal information, which helps the network in localization and enables velocity prediction.
We experiment with using $1$, $5$, and $10$ lidar sweeps.
The results show that both detection and velocity estimates improve with an increasing number of lidar sweeps but with diminishing rate of return (\tableref{table:lidar_exp}).

\begin{table}[]
\setlength\tabcolsep{1.1pt} % Decrease this to reduce margins in each cell.
\footnotesize
\centering
\begin{tabular}{| c | c | c | c | c |}
\hline
\textbf{Lidar sweeps}   & \textbf{Pretraining}	& \textbf{NDS (\%)} & \textbf{mAP (\%)} & \textbf{mAVE (m/s)} \\ \hline
\hline
1					    & KITTI		& 31.8		& 21.9 & 1.21	    \\ \hline
5					    & KITTI		& 42.9		& 27.7 & 0.34	    \\ \hline
10					    & KITTI		& \textbf{44.8}	& 28.8 & 0.30 	\\ \hline
10					    & ImageNet	& \textbf{44.9}	& 28.9 & 0.31	\\ \hline
10					    & None		& 44.2		& 27.6 & 0.33		\\ \hline
\end{tabular}
\vspace{+1mm}
\caption{PointPillars~\cite{pointpillars} detection performance on the val set.
We can see that more lidar sweeps lead to a significant performance increase and that pretraining with ImageNet is on par with KITTI.
}
\vspace{-1mm}
\label{table:lidar_exp}
\end{table}

\mypar{Which sensor is most important?}
An important question for AVs is which sensors are required to achieve the best detection performance.
Here we compare the performance of leading lidar and image detectors.
We focus on these modalities as there are no competitive radar-only methods in the literature and our preliminary study with PointPillars on radar data did not achieve promising results.
We compare PointPillars, which is a fast and light lidar detector with MonoDIS, a top image detector (\tableref{table:detection_challenge}).
The two methods achieve similar mAP (30.5\% vs. 30.4\%), but PointPillars has higher NDS (45.3\% vs. 38.4\%).
The close mAP is, of itself, notable and speaks to the recent advantage in 3D estimation from monocular vision.
However, as discussed above the differences would be larger with an IOU based matching function.

Class specifc performance is in \tableref{table:class_results}-SM.
PointPillars was stronger for the two most common classes: cars ($68.4\%$ vs. $47.8\%$ AP), and pedestrians ($59.7\%$ vs. $37.0\%$ AP).
MonoDIS, on the other hand, was stronger for the smaller classes bicycles ($24.5\%$ vs. $1.1\%$ AP) and cones ($48.7\%$ vs. $30.8\%$ AP).
This is expected since
1) bicycles are thin objects with typically few lidar returns and
2) traffic cones are easy to detect in images, but small and easily overlooked in a lidar pointcloud.
3) MonoDIS applied importance sampling during training to boost rare classes.
With similar detection performance, why was NDS lower for MonoDIS?
The main reasons are the average translation errors ($52$cm vs. $74$cm) and velocity errors ($1.55 m/s$ vs. $0.32 m/s$), both as expected.
MonoDIS also had larger scale errors with mean IOU $74\%$ vs. $71\%$ but the difference is small,
suggesting the strong ability for image-only methods to infer size from appearance.

\begin{table}[]
\setlength\tabcolsep{2.0pt} % Decrease this to reduce margins in each cell.
\footnotesize
\begin{tabular}{| C{1.4cm} | C{0.75cm} | C{0.75cm} | C{0.8cm} | C{0.81cm} | C{0.85cm} | C{0.8cm} | C{0.85cm} |} \hline
\multirow{2}{*}{\textbf{Method}} & \textbf{NDS} & \textbf{mAP} & \textbf{mATE} & \textbf{mASE} & \textbf{mAOE} & \textbf{mAVE} & \textbf{mAAE} \\ \cline{2-8}
\textbf & (\%) & (\%) & (m) & (1-iou) & (rad) & (m/s) & (1-acc) \\ \hline \hline
OFT~\cite{oft}$^{\dag}$             &   21.2    &   12.6    &   0.82  &   0.36  &   0.85  &   1.73  &    0.48   \\ \hline
SSD+3D$^{\dag}$                     &   26.8    &   16.4    &   0.90  &   0.33  &   0.62  &   1.31  &    0.29   \\ \hline
MDIS~\cite{monodis}$^{\dag}$        &   38.4    &   30.4    &   0.74  &   0.26  &   0.55  &   1.55  &    0.13   \\ \hline
PP~\cite{pointpillars}              &   45.3    &   30.5    &   0.52  &   0.29  &   0.50  &   0.32  &    0.37   \\ \hline
Megvii \cite{megvii}                & \textbf{63.3} & \textbf{52.8}   &   \textbf{0.30} &   \textbf{0.25} & \textbf{0.38}   & \textbf{0.25}   & \textbf{0.14}\\ \hline
\end{tabular}
\vspace{+1mm}
\caption{
Object detection results on the test set of nuScenes.
PointPillars, OFT and SSD+3D are baselines provided in this paper, other methods are the top submissions to the nuScenes detection challenge leaderboard.
$({\dag})$ use only monocular camera images as input.
All other methods use lidar. PP: PointPillars~\cite{pointpillars}, MDIS: MonoDIS~\cite{monodis}.
}
\label{table:detection_challenge}
\end{table}

\mypar{The importance of pre-training.}
Using the lidar baseline we examine the importance of pre-training when training a detector on nuScenes.
No pretraining means weights are initialized randomly using a uniform distribution as in~\cite{heInit}.
ImageNet~\cite{imagenet} pretraining~\cite{alexnet} uses a backbone that was first trained to accurately classify images.
KITTI~\cite{kitti} pretraining uses a backbone that was trained on the lidar pointclouds to predict 3D boxes.
Interestingly, while the KITTI pretrained network did converge faster, the final performance of the network only marginally varied between different pretrainings (\tableref{table:lidar_exp}).
One explanation may be that while KITTI is close in domain, the size is not large enough.

\mypar{Better detection gives better tracking.}
Weng and Kitani~\cite{ab3dmot} presented a simple baseline that achieved state-of-the-art 3d tracking results using powerful detections on KITTI.
Here we analyze whether better detections also imply better tracking performance on nuScenes, using the image and lidar baselines presented in \secref{sec:experiments:baselines}.
Megvii, PointPillars and MonoDIS achieve an sAMOTA of $17.9\%$, $3.5\%$ and $4.5\%$, and an AMOTP of $1.50m$, $1.69m$ and $1.79m$ on the val set.
Compared to the mAP and NDS detection results in \tableref{table:detection_challenge}, the ranking is similar.
While the performance is correlated across most metrics, we notice that MonoDIS has the shortest LGD and highest number of track fragmentations.
This may indicate that despite the lower performance, image based methods are less likely to miss an object for a protracted period of time.

% !TEX root = ../nuscenes.tex

\vspace{-2mm}
\squeeze
\section{Conclusion}
\label{sec:conclusion}
\squeeze
\vspace{-2mm}
In this paper we present the nuScenes dataset, detection and tracking tasks, metrics, baselines and results.
This is the first dataset collected from an AV approved for testing on public roads and that contains the full \ang{360} sensor suite (lidar, images, and radar).
nuScenes has the largest collection of 3D box annotations of any previously released dataset.
To spur research on 3D object detection for AVs, we introduce a new detection metric that balances all aspects of detection performance.
We demonstrate novel adaptations of leading lidar and image object detectors and trackers on nuScenes.
Future work will add image-level and point-level semantic labels and a benchmark for trajectory prediction~\cite{covernet}.

\vspace{+1mm}
\mypar{Acknowledgements.}
The nuScenes dataset was annotated by Scale.ai and we thank Alexandr Wang and Dave Morse for their support.
We thank Sun Li, Serene Chen and Karen Ngo at nuTonomy for data inspection and quality control,
Bassam Helou and Thomas Roddick for OFT baseline results,
Sergi Widjaja and Kiwoo Shin for the tutorials,
and Deshraj Yadav and Rishabh Jain from EvalAI~\cite{evalai} for setting up the nuScenes challenges.

{\small
\bibliographystyle{ieee_fullname}
\bibliography{../references}

\begin{thebibliography}{10}\itemsep=-1pt

\bibitem{alessandretti2007vehicle}
Giancarlo Alessandretti, Alberto Broggi, and Pietro Cerri.
\newblock Vehicle and guard rail detection using radar and vision data fusion.
\newblock {\em IEEE Transactions on Intelligent Transportation Systems}, 2007.

\bibitem{mapprior_tl}
Dan Barnes, Will Maddern, and Ingmar Posner.
\newblock Exploiting 3d semantic scene priors for online traffic light
  interpretation.
\newblock In {\em IVS}, 2015.

\bibitem{3decades_adas}
Klaus Bengler, Klaus Dietmayer, Berthold Farber, Markus Maurer, Christoph
  Stiller, and Hermann Winner.
\newblock Three decades of driver assistance systems: Review and future
  perspectives.
\newblock {\em ITSM}, 2014.

\bibitem{motmetrics}
Keni Bernardin, Alexander Elbs, and Rainer Stiefelhagen.
\newblock Multiple object tracking performance metrics and evaluation in a
  smart room environment.
\newblock In {\em ECCV Workshop on Visual Surveillance}, 2006.

\bibitem{monoloco}
Lorenzo Bertoni, Sven Kreiss, and Alexandre Alahi.
\newblock Monoloco: Monocular 3d pedestrian localization and uncertainty
  estimation.
\newblock In {\em ICCV}, 2019.

\bibitem{malaga_urban}
Jos\'{e}-Luis Blanco-Claraco, Francisco-{\'A}ngel Moreno-Dueñas, and Javier
  Gonz{\'a}lez-Jim\'{e}nez.
\newblock The {M{\'a}laga} urban dataset: High-rate stereo and lidar in a
  realistic urban scenario.
\newblock {\em IJRR}, 2014.

\bibitem{kittigps}
Martin Brossard, Axel Barrau, and Silv\`ere Bonnabel.
\newblock {AI-IMU Dead-Reckoning}.
\newblock {\em arXiv preprint arXiv:1904.06064}, 2019.

\bibitem{camvid}
Gabriel~J. Brostow, Jamie Shotton, Julien Fauqueur, and Roberto Cipolla.
\newblock Segmentation and recognition using structure from motion point
  clouds.
\newblock In {\em ECCV}, 2008.

\bibitem{spagnn-uber}
Sergio Casas, Cole Gulino, Renjie Liao, and Raquel Urtasun.
\newblock Spatially-aware graph neural networks for relational behavior
  forecasting from sensor data.
\newblock {\em arXiv preprint arXiv:1910.08233}, 2019.

\bibitem{argoverse}
Ming-Fang Chang, John~W Lambert, Patsorn Sangkloy, Jagjeet Singh, Slawomir Bak,
  Andrew Hartnett, De Wang, Peter Carr, Simon Lucey, Deva Ramanan, and James
  Hays.
\newblock Argoverse: 3d tracking and forecasting with rich maps.
\newblock In {\em CVPR}, 2019.

\bibitem{d2_city}
Z. Che, G. Li, T. Li, B. Jiang, X. Shi, X. Zhang, Y. Lu, G. Wu, Y. Liu, and J.
  Ye.
\newblock {$D^{2}$-City}: A large-scale dashcam video dataset of diverse
  traffic scenarios.
\newblock {\em arXiv:1904.01975}, 2019.

\bibitem{3dop}
Xiaozhi Chen, Kaustav Kundu, Yukun Zhu, Andrew~G Berneshawi, Huimin Ma, Sanja
  Fidler, and Raquel Urtasun.
\newblock 3d object proposals for accurate object class detection.
\newblock In {\em NIPS}, 2015.

\bibitem{mono3d}
Xiaozhi Chen, Laustav Kundu, Ziyu Zhang, Huimin Ma, Sanja Fidler, and Raquel
  Urtasun.
\newblock Monocular 3d object detection for autonomous driving.
\newblock In {\em CVPR}, 2016.

\bibitem{mv3d}
Xiaozhi Chen, Huimin Ma, Ji Wan, Bo Li, and Tian Xia.
\newblock Multi-view 3d object detection network for autonomous driving.
\newblock In {\em CVPR}, 2017.

\bibitem{lidar_video}
Yiping Chen, Jingkang Wang, Jonathan Li, Cewu Lu, Zhipeng Luo, Han Xue, and
  Cheng Wang.
\newblock Lidar-video driving dataset: Learning driving policies effectively.
\newblock In {\em CVPR}, 2018.

\bibitem{stanford2020nuscenes}
Hsu-kuang Chiu, Antonio Prioletti, Jie Li, and Jeannette Bohg.
\newblock Probabilistic 3d multi-object tracking for autonomous driving.
\newblock {\em arXiv preprint arXiv:2001.05673}, 2020.

\bibitem{kaist}
Yukyung Choi, Namil Kim, Soonmin Hwang, Kibaek Park, Jae~Shin Yoon, Kyounghwan
  An, and In~So Kweon.
\newblock {KAIST} multi-spectral day/night data set for autonomous and assisted
  driving.
\newblock {\em IEEE Transactions on Intelligent Transportation Systems}, 2017.

\bibitem{localization}
Z.~J. Chong, B. Qin, T. Bandyopadhyay, M.~H. Ang, E. Frazzoli, and D. Rus.
\newblock Synthetic 2d lidar for precise vehicle localization in 3d urban
  environment.
\newblock In {\em ICRA}, 2013.

\bibitem{cityscapes}
Marius Cordts, Mohamed Omran, Sebastian Ramos, Timo Rehfeld, Markus Enzweiler,
  Rodrigo Benenson, Uwe Franke, Stefan Roth, and Bernt Schiele.
\newblock The {Cityscapes} dataset for semantic urban scene understanding.
\newblock In {\em CVPR}, 2016.

\bibitem{hog}
Navneet Dalal and Bill Triggs.
\newblock Histograms of oriented gradients for human detection.
\newblock In {\em CVPR}, 2005.

\bibitem{imagenet}
Jia Deng, Wei Dong, Richard Socher, Li-Jia Li, Kai Li, and Li Fei-Fei.
\newblock {ImageNet}: A large-scale hierarchical image database.
\newblock In {\em CVPR}, 2009.

\bibitem{talk2car}
Thierry Deruyttere, Simon Vandenhende, Dusan Grujicic, Luc Van~Gool, and
  Marie-Francine Moens.
\newblock Talk2car: Taking control of your self-driving car.
\newblock {\em arXiv preprint arXiv:1909.10838}, 2019.

\bibitem{caltech_ped}
Piotr Doll\'ar, Christian Wojek, Bernt Schiele, and Pietro Perona.
\newblock Pedestrian detection: An evaluation of the state of the art.
\newblock {\em PAMI}, 2012.

\bibitem{daimler_ped}
Markus Enzweiler and Dariu~M. Gavrila.
\newblock Monocular pedestrian detection: Survey and experiments.
\newblock {\em PAMI}, 2009.

\bibitem{eth_persons}
Andreas Ess, Bastian Leibe, Konrad Schindler, and Luc Van~Gool.
\newblock A mobile vision system for robust multi-person tracking.
\newblock In {\em CVPR}, 2008.

\bibitem{pascal}
Mark Everingham, Luc Van~Gool, Christopher K.~I. Williams, John Winn, and
  Andrew Zisserman.
\newblock The pascal visual object classes ({VOC}) challenge.
\newblock {\em International Journal of Computer Vision}, 2010.

\bibitem{pointrnn}
Hehe Fan and Yi Yang.
\newblock {PointRNN}: Point recurrent neural network for moving point cloud
  processing.
\newblock {\em arXiv preprint arXiv:1910.08287}, 2019.

\bibitem{datasetsurvey1}
Di Feng, Christian Haase-Schuetz, Lars Rosenbaum, Heinz Hertlein, Fabian
  Duffhauss, Claudius Glaeser, Werner Wiesbeck, and Klaus Dietmayer.
\newblock Deep multi-modal object detection and semantic segmentation for
  autonomous driving: Datasets, methods, and challenges.
\newblock {\em arXiv preprint arXiv:1902.07830}, 2019.

\bibitem{multimodalsurvey}
D. Feng, C. Haase-Schuetz, L. Rosenbaum, H. Hertlein, C. Glaeser, F. Timm, W.
  Wiesbeck, and K. Dietmayer.
\newblock Deep multi-modal object detection and semantic segmentation for
  autonomous driving: Datasets, methods, and challenges.
\newblock {\em arXiv:1902.07830}, 2019.

\bibitem{evalai}
EvalAI: Towards Better Evaluation~Systems for AI~Agents.
\newblock D. yadav and r. jain and h. agrawal and p. chattopadhyay and t. singh
  and a. jain and s. b. singh and s. lee and d. batra.
\newblock {\em arXiv:1902.03570}, 2019.

\bibitem{streetview}
Andrea Frome, German Cheung, Ahmad Abdulkader, Marco Zennaro, Bo Wu, Alessandro
  Bissacco, Hartwig Adam, Hartmut Neven, and Luc Vincent.
\newblock Large-scale privacy protection in google street view.
\newblock In {\em ICCV}, 2009.

\bibitem{kitti}
Andreas Geiger, Philip Lenz, and Raquel Urtasun.
\newblock Are we ready for autonomous driving? the {KITTI} vision benchmark
  suite.
\newblock In {\em CVPR}, 2012.

\bibitem{mapillary}
Neuhold Gerhard, Tobias Ollmann, Samuel~Rota Bulo, and Peter Kontschieder.
\newblock The {Mapillary Vistas} dataset for semantic understanding of street
  scenes.
\newblock In {\em ICCV}, 2017.

\bibitem{audi_a2d2}
Jakob Geyer, Yohannes Kassahun, Mentar Mahmudi, Xavier Ricou, Rupesh Durgesh,
  Andrew~S. Chung, Lorenz Hauswald, Viet~Hoang Pham, Maximilian Mühlegg,
  Sebastian Dorn, Tiffany Fernandez, Martin Jänicke, Sudesh Mirashi,
  Chiragkumar Savani, Martin Sturm, Oleksandr Vorobiov, and Peter Schuberth.
\newblock {A2D2}: {AEV} autonomous driving dataset.
\newblock \url{http://www.a2d2.audi}, 2019.

\bibitem{semantic_map_parking}
Hugo Grimmett, Mathias Buerki, Lina Paz, Pedro Pinies, Paul Furgale, Ingmar
  Posner, and Paul Newman.
\newblock Integrating metric and semantic maps for vision-only automated
  parking.
\newblock In {\em ICRA}, 2015.

\bibitem{driveability}
Junyao Guo, Unmesh Kurup, and Mohak Shah.
\newblock Is it safe to drive? an overview of factors, challenges, and datasets
  for driveability assessment in autonomous driving.
\newblock {\em arXiv:1811.11277}, 2018.

\bibitem{rain}
Shirsendu~Sukanta Halder, Jean-Francois Lalonde, and Raoul~de Charette.
\newblock Physics-based rendering for improving robustness to rain.
\newblock In {\em ICCV}, 2019.

\bibitem{heInit}
Kaiming He, Xiangyu Zhang, Shaoqing Ren, and Jian Sun.
\newblock Delving deep into rectifiers: Surpassing human-level performance on
  imagenet classification.
\newblock In {\em ICCV}, 2015.

\bibitem{resnet}
Kaiming He, Xiangyu Zhang, Shaoqing Ren, and Jian Sun.
\newblock Deep residual learning for image recognition.
\newblock In {\em CVPR}, 2016.

\bibitem{struc_online_maps}
Namdar Homayounfar, Wei-Chiu Ma, Shrinidhi Kowshika~Lakshmikanth, and Raquel
  Urtasun.
\newblock Hierarchical recurrent attention networks for structured online maps.
\newblock In {\em CVPR}, 2018.

\bibitem{apolloscape}
Xinyu Huang, Peng Wang, Xinjing Cheng, Dingfu Zhou, Qichuan Geng, and Ruigang
  Yang.
\newblock The apolloscape open dataset for autonomous driving and its
  application.
\newblock {\em arXiv:1803.06184}, 2018.

\bibitem{rvnet}
Vijay John and Seiichi Mita.
\newblock Rvnet: Deep sensor fusion of monocular camera and radar for
  image-based obstacle detection in challenging environments, 2019.

\bibitem{3d_outdoor}
Hojung Jung, Yuki Oto, Oscar~M. Mozos, Yumi Iwashita, and Ryo Kurazume.
\newblock Multi-modal panoramic 3d outdoor datasets for place categorization.
\newblock In {\em IROS}, 2016.

\bibitem{kalmanfilter}
Rudolph~Emil Kalman.
\newblock A new approach to linear filtering and prediction problems.
\newblock {\em Transactions of the ASME--Journal of Basic Engineering},
  82(Series D):35--45, 1960.

\bibitem{lyftl5}
R. Kesten, M. Usman, J. Houston, T. Pandya, K. Nadhamuni, A. Ferreira, M. Yuan,
  B. Low, A. Jain, P. Ondruska, S. Omari, S. Shah, A. Kulkarni, A. Kazakova, C.
  Tao, L. Platinsky, W. Jiang, and V. Shet.
\newblock {Lyft Level 5 AV Dataset 2019}.
\newblock \url{https://level5.lyft.com/dataset/}, 2019.

\bibitem{sensorblocking}
Jaekyum Kim, Jaehyung Choi, Yechol Kim, Junho Koh, Chung~Choo Chung, and
  Jun~Won Choi.
\newblock Robust camera lidar sensor fusion via deep gated information fusion
  network.
\newblock In {\em IVS}, 2018.

\bibitem{alexnet}
Alex Krizhevsky, Ilya Sutskever, and Geoffrey~E Hinton.
\newblock Imagenet classification with deep convolutional neural networks.
\newblock In {\em NIPS}, 2012.

\bibitem{avod}
Jason Ku, Melissa Mozifian, Jungwook Lee, Ali Harakeh, and Steven Waslander.
\newblock Joint 3d proposal generation and object detection from view
  aggregation.
\newblock In {\em IROS}, 2018.

\bibitem{datasetsurvey2}
Charles-{\'E}ric~No{\"e}l Laflamme, Fran{\c{c}}ois Pomerleau, and Philippe
  Gigu{\`e}re.
\newblock Driving datasets literature review.
\newblock {\em arXiv preprint arXiv:1910.11968}, 2019.

\bibitem{autonue}
Nitheesh Lakshminarayana.
\newblock Large scale multimodal data capture, evaluation and maintenance
  framework for autonomous driving datasets.
\newblock In {\em ICCVW}, 2019.

\bibitem{pointpillars}
Alex~H. Lang, Sourabh Vora, Holger Caesar, Lubing Zhou, Jiong Yang, and Oscar
  Beijbom.
\newblock Pointpillars: Fast encoders for object detection from point clouds.
\newblock In {\em CVPR}, 2019.

\bibitem{contfuse}
Ming Liang, Bin Yang, Shenlong Wang, and Raquel Urtasun.
\newblock Deep continuous fusion for multi-sensor 3d object detection.
\newblock In {\em ECCV}, 2018.

\bibitem{ssd}
Wei Liu, Dragomir Anguelov, Dumitru Erhan, Christian Szegedy, Scott Reed,
  Cheng-Yang Fu, and Alexander~C Berg.
\newblock {SSD}: Single shot multibox detector.
\newblock In {\em ECCV}, 2016.

\bibitem{traffic_predict}
Yuexin Ma, Xinge Zhu, Sibo Zhang, Ruigang Yang, Wenping Wang, and Dinesh
  Manocha.
\newblock Trafficpredict: Trajectory prediction for heterogeneous
  traffic-agents \url{http://apolloscape.auto/tracking.html}.
\newblock In {\em AAAI}, 2019.

\bibitem{robotcar}
Will Maddern, Geoffrey Pascoe, Chris Linegar, and Paul Newman.
\newblock 1 year, 1000 km: The oxford robotcar dataset.
\newblock {\em IJRR}, 2017.

\bibitem{lasernet}
Gregory~P Meyer, Ankit Laddha, Eric Kee, Carlos Vallespi-Gonzalez, and Carl~K
  Wellington.
\newblock Lasernet: An efficient probabilistic 3d object detector for
  autonomous driving.
\newblock In {\em CVPR}, 2019.

\bibitem{deep3dbox}
Arsalan Mousavian, Dragomir Anguelov, John Flynn, and Jana Kosecka.
\newblock 3d bounding box estimation using deep learning and geometry.
\newblock In {\em CVPR}, 2017.

\bibitem{nightowl}
Lukáš Neumann, Michelle Karg, Shanshan Zhang, Christian Scharfenberger, Eric
  Piegert, Sarah Mistr, Olga Prokofyeva, Robert Thiel, Andrea Vedaldi, Andrew
  Zisserman, and Bernt Schiele.
\newblock Nightowls: A pedestrians at night dataset.
\newblock In {\em ACCV}, 2018.

\bibitem{starnet}
Jiquan Ngiam, Benjamin Caine, Wei Han, Brandon Yang, Yuning Chai, Pei Sun, Yin
  Zhou, Xi Yi, Ouais Alsharif, Patrick Nguyen, Zhifeng Chen, Jonathon Shlens,
  and Vijay Vasudevan.
\newblock Starnet: Targeted computation for object detection in point clouds.
\newblock {\em arXiv preprint arXiv:1908.11069}, 2019.

\bibitem{howmuchrealdata}
Farzan~Erlik Nowruzi, Prince Kapoor, Dhanvin Kolhatkar, Fahed~Al Hassanat,
  Robert Laganiere, and Julien Rebut.
\newblock How much real data do we actually need: Analyzing object detection
  performance using synthetic and real data.
\newblock In {\em ICML Workshop on AI for Autonomous Driving}, 2019.

\bibitem{h3d}
Abhishek Patil, Srikanth Malla, Haiming Gang, and Yi-Ting Chen.
\newblock The {H3D} dataset for full-surround 3d multi-object detection and
  tracking in crowded urban scenes.
\newblock In {\em ICRA}, 2019.

\bibitem{astar3d}
Quang-Hieu Pham, Pierre Sevestre, Ramanpreet~Singh Pahwa, Huijing Zhan, Chun~Ho
  Pang, Yuda Chen, Armin Mustafa, Vijay Chandrasekhar, and Jie Lin.
\newblock {A*3D Dataset}: Towards autonomous driving in challenging
  environments.
\newblock {\em arXiv:1909.07541}, 2019.

\bibitem{covernet}
Tung Phan-Minh, Elena~Corina Grigore, Freddy~A. Boulton, Oscar Beijbom, and
  Eric~M. Wolff.
\newblock Covernet: Multimodal behavior prediction using trajectory sets.
\newblock In {\em CVPR}, 2020.

\bibitem{frustum}
Charles~R Qi, Wei Liu, Chenxia Wu, Hao Su, and Leonidas~J. Guibas.
\newblock Frustum pointnets for 3d object detection from {RGB-D} data.
\newblock In {\em CVPR}, 2018.

\bibitem{hondadataset}
Vasili Ramanishka, Yi-Ting Chen, Teruhisa Misu, and Kate Saenko.
\newblock Toward driving scene understanding: A dataset for learning driver
  behavior and causal reasoning.
\newblock In {\em CVPR}, 2018.

\bibitem{gpp}
Akshay Rangesh and Mohan~M. Trivedi.
\newblock Ground plane polling for 6dof pose estimation of objects on the road.
\newblock In {\em arXiv:1811.06666}, 2018.

\bibitem{fasterrcnn}
Shaoqing Ren, Kaiming He, Ross Girshick, and Jian Sun.
\newblock Faster {R-CNN}: Towards real-time object detection with region
  proposal networks.
\newblock In {\em NIPS}, 2015.

\bibitem{precog}
Nicholas Rhinehart, Rowan McAllister, Kris~M. Kitani, and Sergey Levine.
\newblock {PRECOG}: Predictions conditioned on goals in visual multi-agent
  scenarios.
\newblock In {\em ICCV}, 2019.

\bibitem{oft}
Thomas Roddick, Alex Kendall, and Roberto Cipolla.
\newblock Orthographic feature transform for monocular 3d object detection.
\newblock In {\em BMVC}, 2019.

\bibitem{monodis}
Andrea Simonelli, Samuel Rota~Bulo, Lorenzo Porzi, Manuel Lopez-Antequera, and
  Peter Kontschieder.
\newblock Disentangling monocular 3d object detection.
\newblock {\em ICCV}, 2019.

\bibitem{onecycle}
Leslie~N. Smith.
\newblock A disciplined approach to neural network hyper-parameters: Part 1 --
  learning rate, batch size, momentum, and weight decay.
\newblock {\em arXiv preprint arXiv:1803.09820}, 2018.

\bibitem{pointpainting}
Sourabh Vora, Alex~H Lang, Bassam Helou, and Oscar Beijbom.
\newblock Pointpainting: Sequential fusion for 3d object detection.
\newblock In {\em CVPR}, 2020.

\bibitem{pseudo_lidar}
Yan Wang, Wei-Lun Chao, Divyansh Garg, Bharath Hariharan, Mark Campbell, and
  Kilian~Q. Weinberger.
\newblock Pseudo-lidar from visual depth estimation: Bridging the gap in 3d
  object detection for autonomous driving.
\newblock In {\em CVPR}, 2019.

\bibitem{scene-understanding}
Ziyan Wang, Buyu Liu, Samuel Schulter, and Manmohan Chandraker.
\newblock Dataset for high-level 3d scene understanding of complex road scenes
  in the top-view.
\newblock In {\em CVPRW}, 2019.

\bibitem{sparsepool}
Zining Wang, Wei Zhan, and Masayoshi Tomizuka.
\newblock Fusing bird's eye view lidar point cloud and front view camera image
  for 3d object detection.
\newblock In {\em IVS}, 2018.

\bibitem{waymo_open_dataset}
Waymo.
\newblock {Waymo Open Dataset}: An autonomous driving dataset, 2019.

\bibitem{ab3dmot}
Xinshuo Weng and Kris Kitani.
\newblock A baseline for 3d multi-object tracking.
\newblock {\em arXiv preprint arXiv:1907.03961}, 2019.

\bibitem{changelives}
L. Woensel and G. Archer.
\newblock Ten technologies which could change our lives.
\newblock {\em European Parlimentary Research Service}, 2015.

\bibitem{tud_brussels_ped}
Christian Wojek, Stefan Walk, and Bernt Schiele.
\newblock Multi-cue onboard pedestrian detection.
\newblock In {\em CVPR}, 2009.

\bibitem{mlf_mono}
Bin Xu and Zhenzhong Chen.
\newblock Multi-level fusion based 3d object detection from monocular images.
\newblock In {\em CVPR}, 2018.

\bibitem{pointfusion}
Danfei Xu, Dragomir Anguelov, and Ashesh Jain.
\newblock Pointfusion: Deep sensor fusion for 3d bounding box estimation.
\newblock In {\em CVPR}, 2018.

\bibitem{hdnet}
Bin Yang, Ming Liang, and Raquel Urtasun.
\newblock {HDNET}: Exploiting {HD} maps for 3d object detection.
\newblock In {\em CoRL}, 2018.

\bibitem{sarpnet}
Yangyang Ye, Chi Zhang, Xiaoli Hao, Houjin Chen, and Zhaoxiang Zhang.
\newblock {SARPNET}: Shape attention regional proposal network for lidar-based
  3d object detection.
\newblock {\em Neurocomputing}, 2019.

\bibitem{woodscape}
Senthil Yogamani, Ciar{\'a}n Hughes, Jonathan Horgan, Ganesh Sistu, Padraig
  Varley, Derek O'Dea, Michal Uric{\'a}r, Stefan Milz, Martin Simon, Karl
  Amende, et~al.
\newblock Woodscape: A multi-task, multi-camera fisheye dataset for autonomous
  driving.
\newblock In {\em ICCV}, 2019.

\bibitem{deepdrive}
Fisher Yu, Wenqi Xian, Yingying Chen, Fangchen Liu, Mike Liao, Vashisht
  Madhavan, and Trevor Darrell.
\newblock {BDD100K}: A diverse driving video database with scalable annotation
  tooling.
\newblock {\em arXiv:1805.04687}, 2018.

\bibitem{datasetsurvey3}
Ekim Yurtsever, Jacob Lambert, Alexander Carballo, and Kazuya Takeda.
\newblock A survey of autonomous driving: Common practices and emerging
  technologies.
\newblock {\em arXiv preprint arXiv:1906.05113}, 2019.

\bibitem{jointface}
Kaipeng Zhang, Zhanpeng Zhang, Zhifeng Li, and Yu Qiao.
\newblock Joint face detection and alignment using multitask cascaded
  convolutional networks.
\newblock {\em SPL}, 23(10), 2016.

\bibitem{citypersons}
Shanshan Zhang, Rodrigo Benenson, and Bernt Schiele.
\newblock Citypersons: A diverse dataset for pedestrian detection.
\newblock In {\em CVPR}, 2017.

\bibitem{datasetsurvey4}
Hao Zhou and Jorge Laval.
\newblock Longitudinal motion planning for autonomous vehicles and its impact
  on congestion: A survey.
\newblock {\em arXiv preprint arXiv:1910.06070}, 2019.

\bibitem{megvii}
Benjin {Zhu}, Zhengkai {Jiang}, Xiangxin {Zhou}, Zeming {Li}, and Gang {Yu}.
\newblock Class-balanced grouping and sampling for point cloud 3d object
  detection.
\newblock {\em arXiv:1908.09492}, 2019.

\bibitem{objspecificdistance}
Jing Zhu and Yi Fang.
\newblock Learning object-specific distance from a monocular image.
\newblock In {\em ICCV}, 2019.

\end{thebibliography}
}
\clearpage
% !TEX root = ../nuscenes_supp.tex

\appendix

\begin{minipage}[t]{1\textwidth}
\centering
\Large \bf \title \par
nuScenes: A multimodal dataset for autonomous driving\\
Supplementary Material
\end{minipage}

%%%%%%%%% TITLE
%\squeeze
%\vspace{+2mm}
\section{The nuScenes dataset}
\squeeze

In this section we provide more details on the nuScenes dataset, the sensor calibration, privacy protection approach, data format, class mapping and annotation statistics.

\mypar{Sensor calibration.}

To achieve a high quality multi-sensor dataset, careful calibration of sensor intrinsic and extrinsic parameters is required.
These calibration parameters are updated around twice per week over the data collection period of 6 months.
Here we describe how we perform sensor calibration for our data collection platform to achieve a high-quality multimodal dataset.
Specifically, we carefully calibrate the extrinsics and intrinsics of every sensor.
We express extrinsic coordinates of each sensor to be relative to the \emph{ego frame},
i.e. the midpoint of the rear vehicle axle.
The most relevant steps are described below:
\begin{itemize} \itemsep 0em
\vspace{-2mm}
\item Lidar extrinsics:
We use a laser liner to accurately measure the relative location of the lidar to the ego frame.
\item Camera extrinsics:
We place a cube-shaped calibration target in front of the camera and lidar sensors.
The calibration target consists of three orthogonal planes with known patterns.
After detecting the patterns we compute the transformation matrix from camera to lidar by aligning the planes of the calibration target.
Given the lidar to ego frame transformation computed above,
we compute the camera to ego frame transformation.
\item Radar extrinsics:
We mount the radar in a horizontal position.
Then we collect radar measurements by driving on public roads.
After filtering radar returns for moving objects,
we calibrate the yaw angle using a brute force approach to minimize the compensated range rates for static objects.
\item Camera intrinsic calibration:
We use a calibration target board with a known set of patterns to infer the intrinsic and distortion parameters of the camera.
\vspace{-3mm}
\end{itemize}

\mypar{Privacy protection.}
It is our priority to protect the privacy of third parties.
As manual labeling of faces and license plates is prohibitively expensive for 1.4M images,
we use state-of-the-art object detection techniques.
Specifically for plate detection, we use Faster R-CNN~\cite{fasterrcnn} with ResNet-101 backbone~\cite{resnet} trained on Cityscapes~\cite{cityscapes}\footnote{https://github.com/bourdakos1/Custom-Object-Detection}. For face detection, we use~\cite{jointface}\footnote{https://github.com/TropComplique/mtcnn-pytorch}.
We set the classification threshold to achieve an extremely high recall (similar to~\cite{streetview}).
To increase the precision, we remove predictions that do not overlap with the reprojections of the known \emph{pedestrian} and \emph{vehicle} boxes in the image.
Eventually we use the predicted boxes to blur faces and license plates in the images.

\mypar{Data format.}
Contrary to most existing datasets~\cite{kitti,h3d, apolloscape}, we store the annotations and metadata (e.g. localization, timestamps, calibration data) in a relational database which avoids redundancy and allows for efficient access.
The nuScenes devkit, taxonomy and annotation instructions are available online\footnote{\url{https://github.com/nutonomy/nuscenes-devkit}}.

\mypar{Class mapping.}
The nuScenes dataset comes with annotations for 23 classes.
Since some of these only have a handful of annotations, we merge similar classes and remove classes that have less than 10000 annotations.
This results in 10 classes for our detection task.
Out of these, we omit 3 classes that are mostly static for the tracking task.
\tableref{table:class_map}-SM shows the detection classes and tracking classes and their counterpart in the general nuScenes dataset.

\begin{table}[]
\vspace{+24.7mm}
\setlength\tabcolsep{2.0pt} % Decrease this to reduce margins in each cell.
\footnotesize
\centering
\begin{tabular}{| c | C{2.8cm} | c | }
\hline
\textbf{General nuScenes class} &  \textbf{Detection class} &   \textbf{Tracking class}	\\ \hline
\hline
animal                    & void  			        &    void                   \\ \hline
debris                    & void			        &    void                   \\ \hline
pushable\_pullable        & void			        &    void                   \\ \hline
bicycle\_rack             & void			        &    void                   \\ \hline
ambulance                 & void			        &    void                   \\ \hline
police                    & void			        &    void                   \\ \hline
barrier                   & barrier				    &    void                   \\ \hline
bicycle                   & bicycle				    &    bicycle                \\ \hline
bus.bendy                 & bus				        &    bus                    \\ \hline
bus.rigid                 & bus				        &    bus                    \\ \hline
car                       & car					    &    car                    \\ \hline
construction              & construction\_vehicle	&    void                   \\ \hline
motorcycle                & motorcycle			    &    motorcycle             \\ \hline
adult                     & pedestrian			    &    pedestrian             \\ \hline
child                     & pedestrian			    &    pedestrian             \\ \hline
construction\_worker      & pedestrian			    &    pedestrian             \\ \hline
police\_officer           & pedestrian			    &    pedestrian             \\ \hline
personal\_mobility        & void			        &    void                   \\ \hline
stroller                  & void			        &    void                   \\ \hline
wheelchair                & void			        &    void                   \\ \hline
trafficcone               & traffic\_ cone			&    void                   \\ \hline
trailer                   & trailer				    &    trailer                \\ \hline
truck                     & truck				    &    truck                  \\ \hline
\end{tabular}
\vspace{+1mm}
\caption{Mapping from general classes in nuScenes to the classes used in the detection and tracking challenges.
Note that for brevity we omit most prefixes for the general nuScenes classes.
}
\label{table:class_map}
\end{table}

\begin{figure}
\begin{center}
\includegraphics[width=\linewidth]{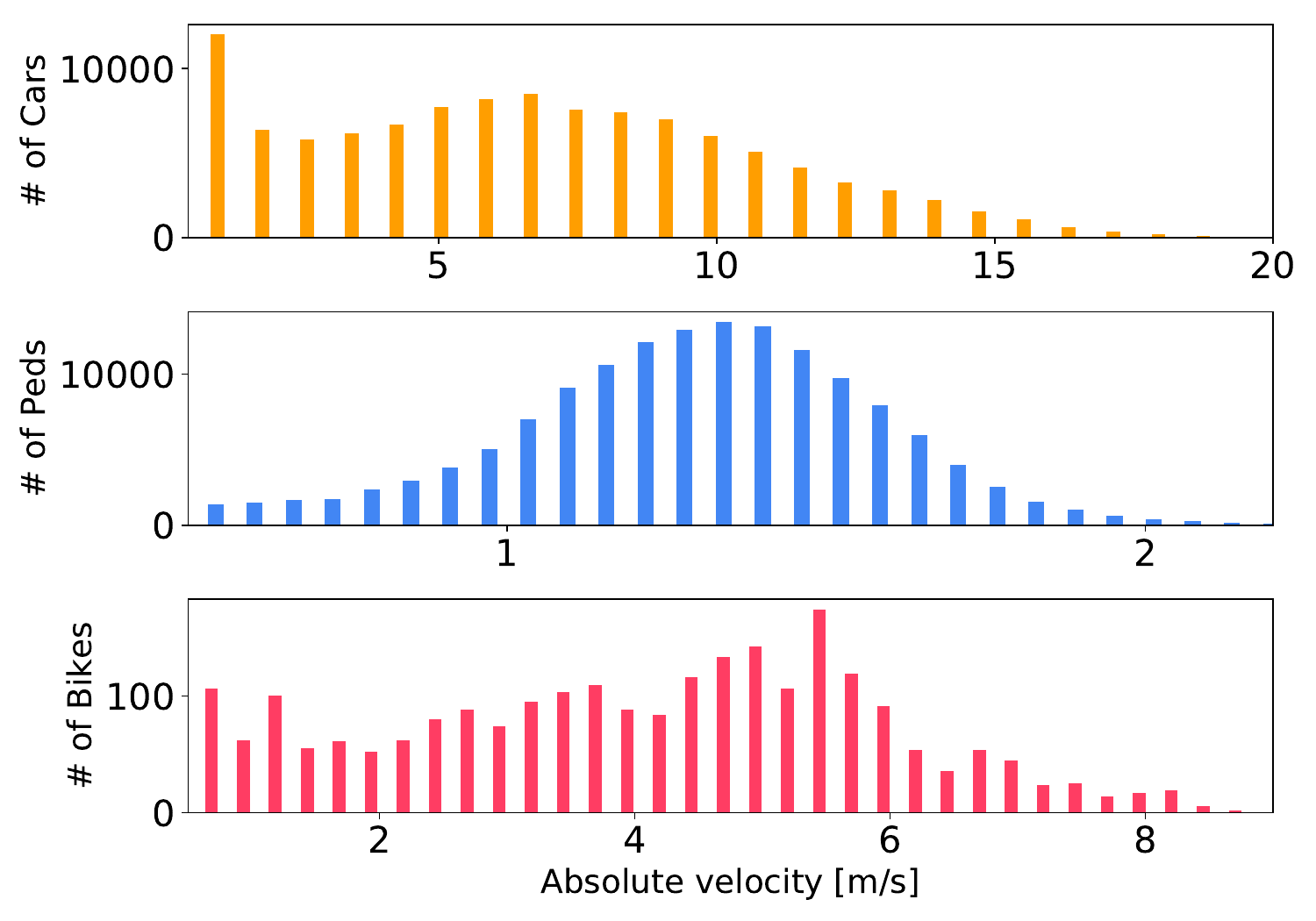}
\end{center}
\squeeze
\vspace{-3mm}
\caption{Absolute velocities. We only look at moving objects with speed $>$ 0.5m/s.}
\vspace{-2mm}
\label{fig:abs_vel}
\end{figure}

\mypar{Annotation statistics.}
We present more statistics on the annotations of nuScenes.
Absolute velocities are shown in \figref{fig:abs_vel}-SM.
The average speed for moving  \emph{car}, \emph{pedestrian} and  \emph{bicycle} categories are $6.6$, $1.3$ and $4$ m/s.
Note that our data was gathered from urban areas which shows reasonable velocity range for these three categories.

\begin{figure}
\begin{center}
\vspace{+1mm}
\includegraphics[width=\linewidth]{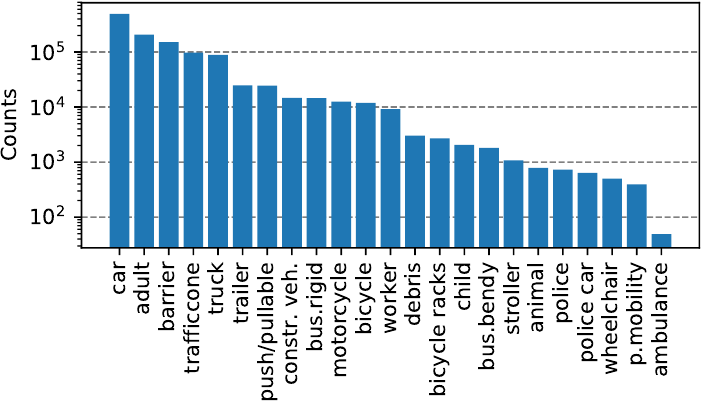}
\vspace{+1mm}
\includegraphics[width=\linewidth, height=4.3cm]{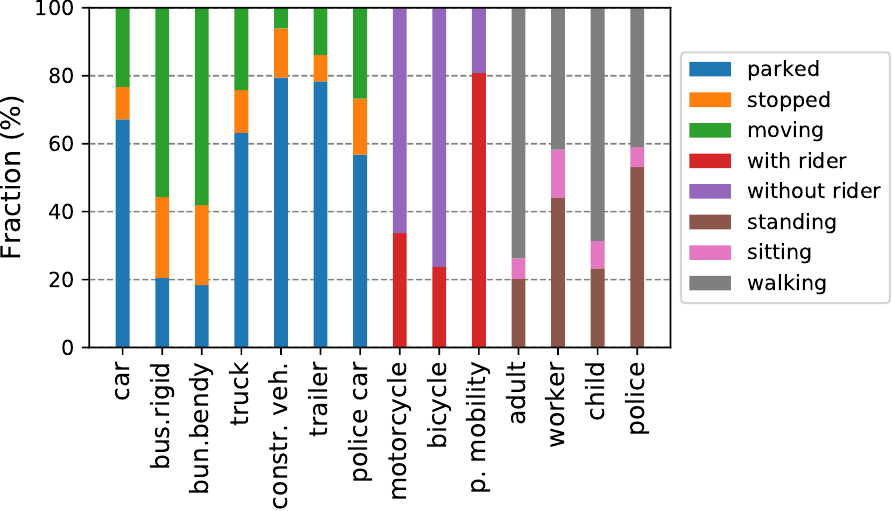}
\end{center}
\squeeze
\vspace{-10pt}
\caption{
Top: Number of annotations per category.
Bottom: Attributes distribution for selected categories.
Cars and adults are the most frequent categories in our dataset, while ambulance is the least frequent.
The attribute plot also shows some expected patterns: construction vehicles are rarely moving, pedestrians are rarely sitting while buses are commonly moving.
}
\label{fig:annotation_count}
\end{figure}

\begin{figure}
\begin{center}
\vspace{-3mm}
\includegraphics[width=\linewidth]{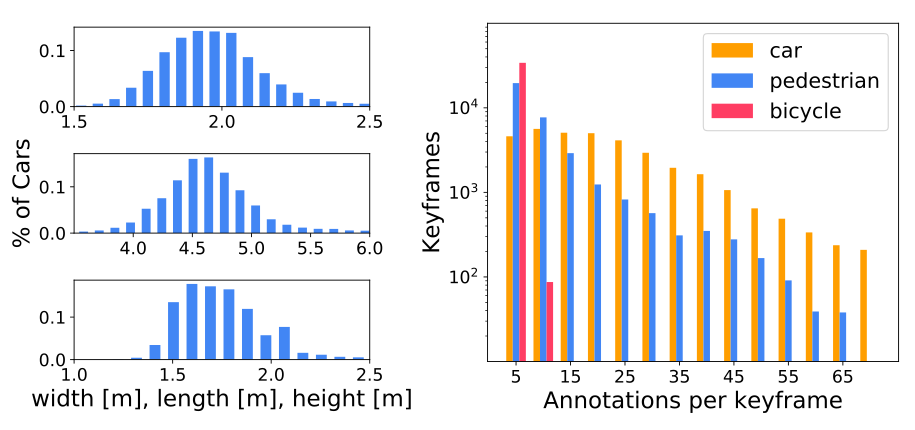}
\end{center}
\squeeze
\vspace{-10pt}
\caption{Left: Bounding box size distributions for \emph{car}. Right: Category count in each keyframe for \emph{car}, \emph{pedestrian}, and \emph{bicycle}.
}
\vspace{-3mm}
\label{fig:basic_stats}
\end{figure}

\begin{figure}
\begin{center}
\includegraphics[width=\linewidth]{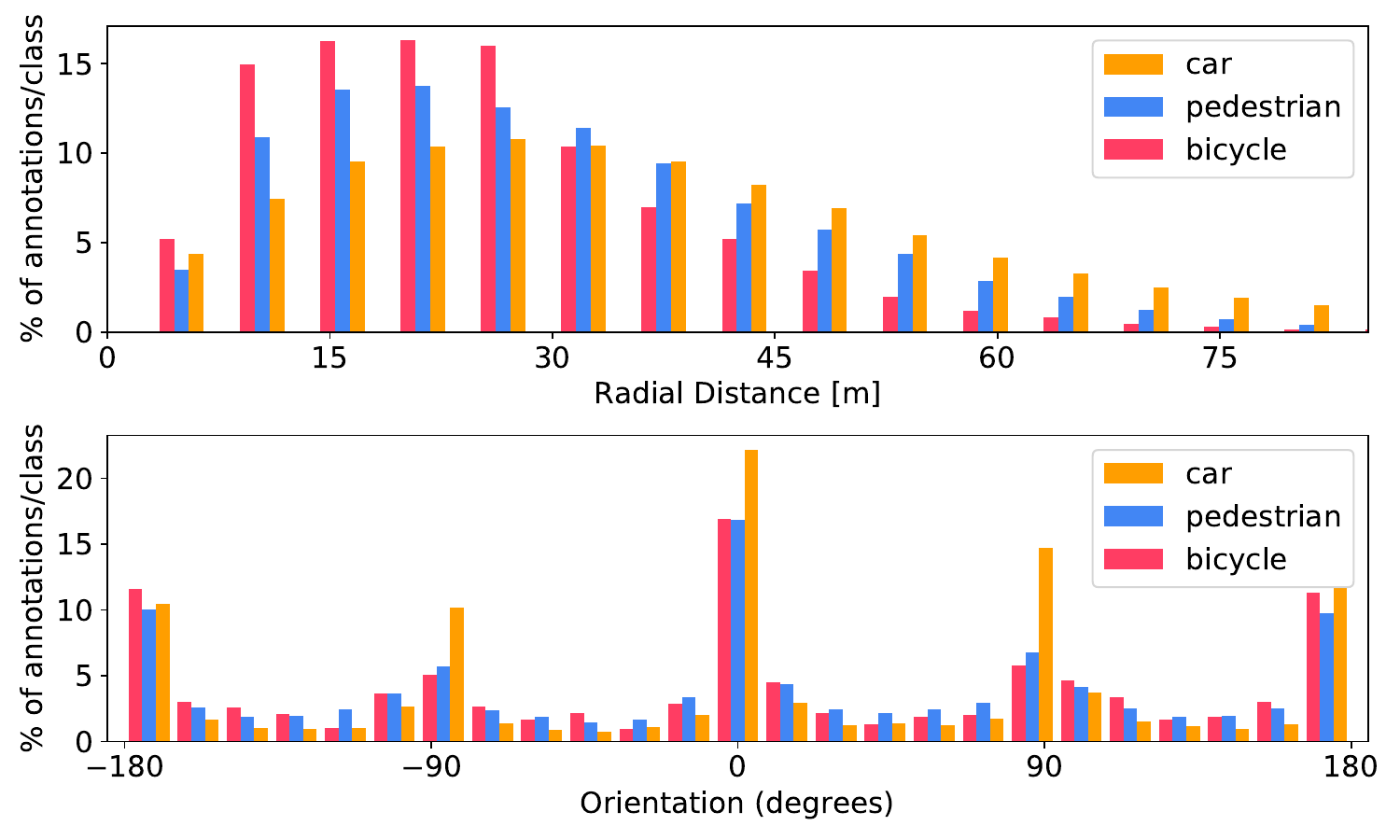}
\end{center}
\squeeze
\vspace{-12pt}
\caption{Top: radial distance of objects from the ego vehicle.
Bottom: orientation of boxes in box coordinate frame.
}
\label{fig:dist_orient_hists}
\end{figure}

\begin{figure*}
\begin{center}
\includegraphics[width=\linewidth]{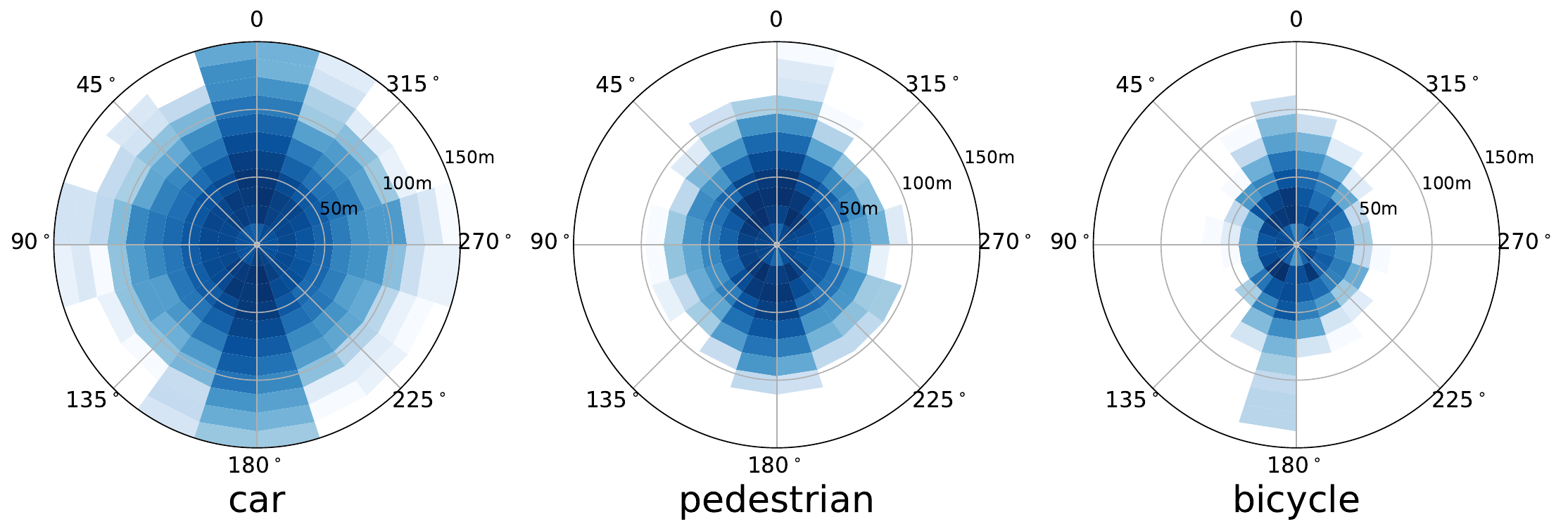}
\end{center}
\squeeze
\vspace{-3mm}
\caption{Polar log-scaled density map for box annotations where the radial axis is the distance from the ego-vehicle in meters and the polar axis is the yaw angle wrt to the ego-vehicle.
The darker the bin is, the more box annotations in that area.
Here, we only show the density up to 150m radial distance for all maps, but~\emph{car} would have annotations up to 200m.}
\vspace{-2mm}
\label{fig:density_yaw}
\end{figure*}

\begin{figure*}
\begin{center}
\includegraphics[width=\linewidth]{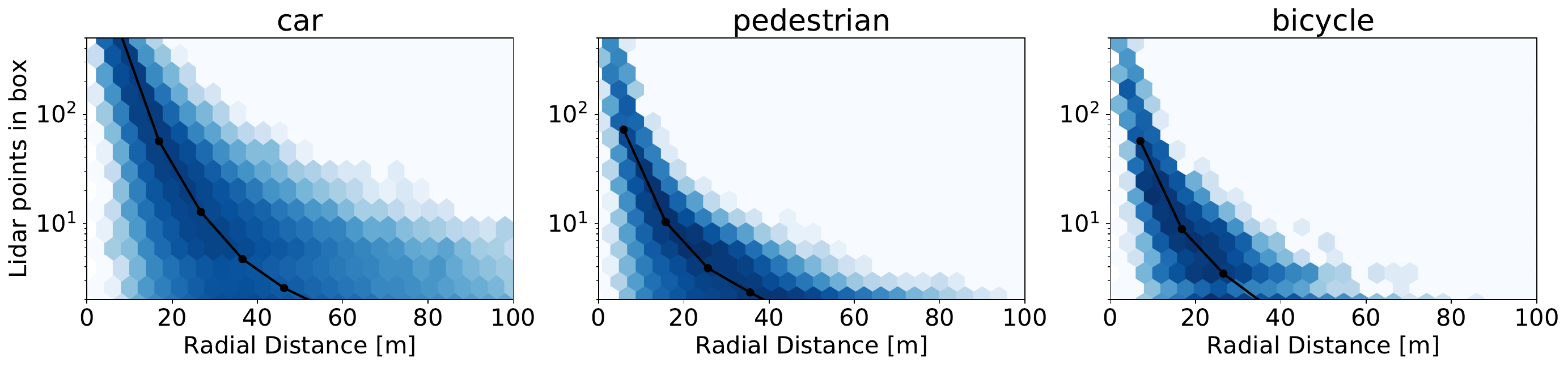}
\end{center}
\squeeze
\vspace{-3mm}
\caption{Hexbin log-scaled density plots of the number of lidar points inside a box annotation stratified by categories (\emph{car}, \emph{pedestrian} and \emph{bicycle}.}
\vspace{+1mm}
\label{fig:lidar_per_cat}
\end{figure*}

\begin{figure}
\begin{center}
\includegraphics[width=\linewidth]{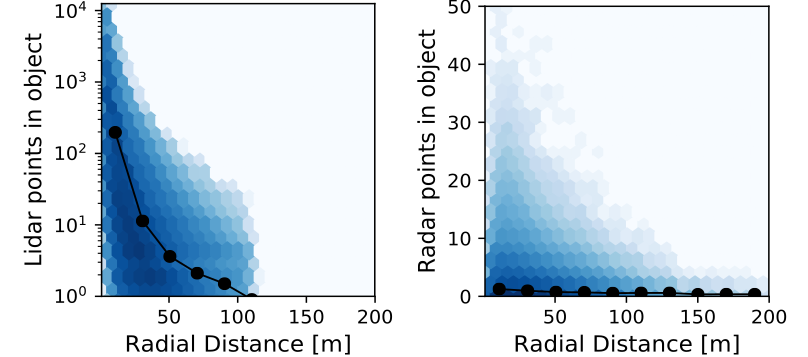}
\end{center}
\squeeze
\vspace{-2mm}
\caption{Hexbin log-scaled density plots of the number of lidar and radar points inside a box annotation.
The black line represents the mean number of points for a given distance wrt the ego-vehicle.}
\vspace{-2mm}
\label{fig:lidar_pt_hists}
\end{figure}

We analyze the distribution of box annotations around the ego-vehicle for \emph{car}, \emph{pedestrian} and \emph{bicycle} categories through a polar range density map as shown in \figref{fig:density_yaw}-SM. Here, the occurrence bins are log-scaled.
Generally, the annotations are well-distributed surrounding the ego-vehicle.
The annotations are also denser when they are nearer to the ego-vehicle.
However, the \emph{pedestrian} and \emph{bicycle} have less annotations above the 100m range.
It can also be seen that the \emph{car} category is denser in the front and back of the ego-vehicle, since most vehicles are following the same lane as the ego-vehicle.

In \secref{sec:dataset} we discussed the number of lidar points inside a box for all categories through a hexbin density plot,
but here we present the number of lidar points of each category as shown in \figref{fig:lidar_per_cat}-SM.
Similarly, the occurrence bins are log-scaled.
As can be seen, there are more lidar points found inside the box annotations for \emph{car} at varying distances from the ego-vehicle as compared to \emph{pedestrian} and \emph{bicycle}.
This is expected as cars have larger and more reflective surface area than the other two categories, hence more lidar points are reflected back to the sensor.

\begin{figure}[H]
\begin{center}
\includegraphics[width=\linewidth]{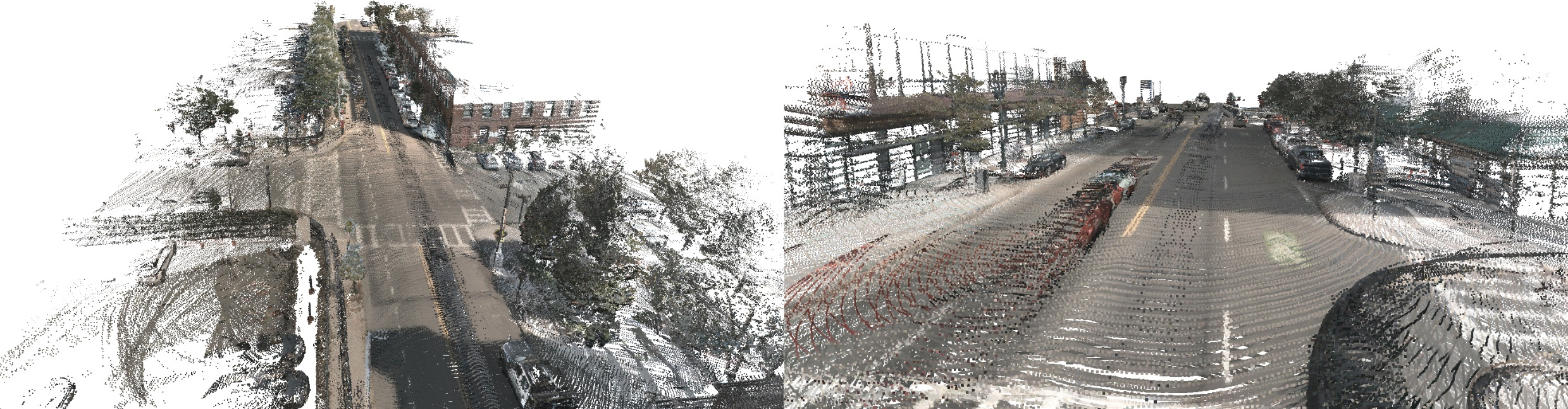}
\end{center}
\squeeze
\vspace{-1mm}
\caption{Sample scene reconstruction given lidar points and camera images. We project the lidar points in an image plane with colors assigned based on the pixel color from the camera data.}
\label{fig:reconstruction}
\end{figure}

\mypar{Scene reconstruction.}
nuScenes uses an accurate lidar based localization algorithm (Section~\ref{sec:dataset}).
It is however difficult to quantify the localization quality, as we do not have ground truth localization data and generally cannot perform loop closure in our scenes.
To analyze our localization qualitatively, we compute the merged pointcloud of an entire scene by registering approximately 800 pointclouds in global coordinates.
We remove points corresponding to the ego vehicle and assign to each point the mean color value of the closest camera pixel that the point is reprojected to.
The result of the scene reconstruction can be seen in \figref{fig:reconstruction}, which demonstrates accurate synchronization and localization.

\squeeze
\section{Implementation details}
\squeeze
Here we provide additional details on training the lidar and image based 3D object detection baselines.

\mypar{PointPillars implementation details.}
For all experiments, our PointPillars~\cite{pointpillars} networks were trained using a pillar xy resolution of 0.25 meters and an x and y range of $[-50, 50]$ meters.
The max number of pillars and batch size was varied with the number of lidar sweeps. For 1, 5, and 10 sweeps, we set the maximum number of pillars to 10000, 22000, and 30000 respectively and the batch size to 64, 64, and 48.
All experiments were trained for 750 epochs.
The initial learning rate was set to $10^{-3}$ and was reduced by a factor of $10$ at epoch 600 and again at 700.
Only ground truth annotations with one or more lidar points in the accumulated pointcloud were used as positive training examples.
Since bikes inside of bike racks are not annotated individually and the evaluation metrics ignore bike racks, all lidar points inside bike racks were filtered out during training.

\mypar{OFT implementation details.}
For each camera, the Orthographic Feature Transform~\cite{oft} (OFT) baseline was trained on a voxel grid in each camera's frame with an lateral range of $[-40, 40]$ meters, a longitudinal range of $[0.1, 50.1]$ meters and a vertical range of $(-3, 1)$ meters.
We trained only on annotations that were within 50 meters of the car's ego frame coordinate system's origin.
Using the `visibility' attribute in the nuScenes dataset, we also filtered out annotations that had visibility less than $40\%$.
The network was trained for 60 epochs using a learning rate of $2 \times 10^{-3}$ and used random initialization for the network weights (no ImageNet pretraining).

\squeeze
\section{Experiments}
\squeeze
In this section we present more detailed result analysis on nuScenes.
We look at the performance on rain and night data, per-class performance and semantic map filtering.
We also analyze the results of the tracking challenge.

\mypar{Performance on rain and night data.}
As described in \secref{sec:dataset}, nuScenes contains data from 2 countries, as well as rain and night data.
The dataset splits (train, val, test) follow the same data distribution with respect to these criteria.
In \tableref{table:rainnight} we analyze the performance of three object detection baselines on the relevant subset of the val set.
We can see a small performance drop for Singapore as compared to the overall val set (USA and Singapore), particularly for vision based methods.
This is likely due to different object appearance in the different countries, as well as different label distributions.
For rain data we see only a small decrease in performance on average, with worse performance for OFT and PP, and slightly better performance for MDIS.
One reason is that the nuScenes dataset annotates any scene with raindrops on the windshield as rainy, regardless of whether there is ongoing rainfall.
Finally, night data shows a drastic performance relative drop of 36\% for the lidar based method and 55\% and 58\% for the vision based methods.
This may indicate that vision based methods are more affected by worse lighting.
We also note that night scenes have very few objects and it is harder to annotate objects with bad visibility.
For annotating data, it is essential to use camera \emph{and} lidar data, as described in \secref{sec:dataset}.

\begin{table}[]
\footnotesize
\begin{tabular}{| C{2.0cm} | C{1.5cm} | C{1.5cm} | C{1.5cm} |} \hline
\textbf{Method}               & \textbf{Singapore} & \textbf{Rain} & \textbf{Night} \\ \hline \hline
OFT~\cite{oft}$^{\dag}$       &    6\%   &  10\%     &    \textbf{55\%}    \\ \hline
MDIS~\cite{monodis}$^{\dag}$  &    8\%   &  -3\%     &    \textbf{58\%}    \\ \hline
PP~\cite{pointpillars}        &    1\%   &   6\%     &    \textbf{36\%}    \\ \hline
\end{tabular}
\caption{
Object detection performance drop evaluated on subsets of the nuScenes val set.
Performance is reported as the relative drop in mAP compared to evaluating on the entire val set.
We evaluate the performance on Singapore data, rain data and night data for three object detection methods.
Note that the MDIS results are not directly comparable to other sections of this work, as a ResNet34~\cite{resnet} backbone and a different training protocol are used.
$({\dag})$ use only monocular camera images as input. PP uses only lidar.
}
\label{table:rainnight}
\end{table}

\begin{figure}
\begin{center}
\vspace{-6mm}
\includegraphics[width=1.05\linewidth]{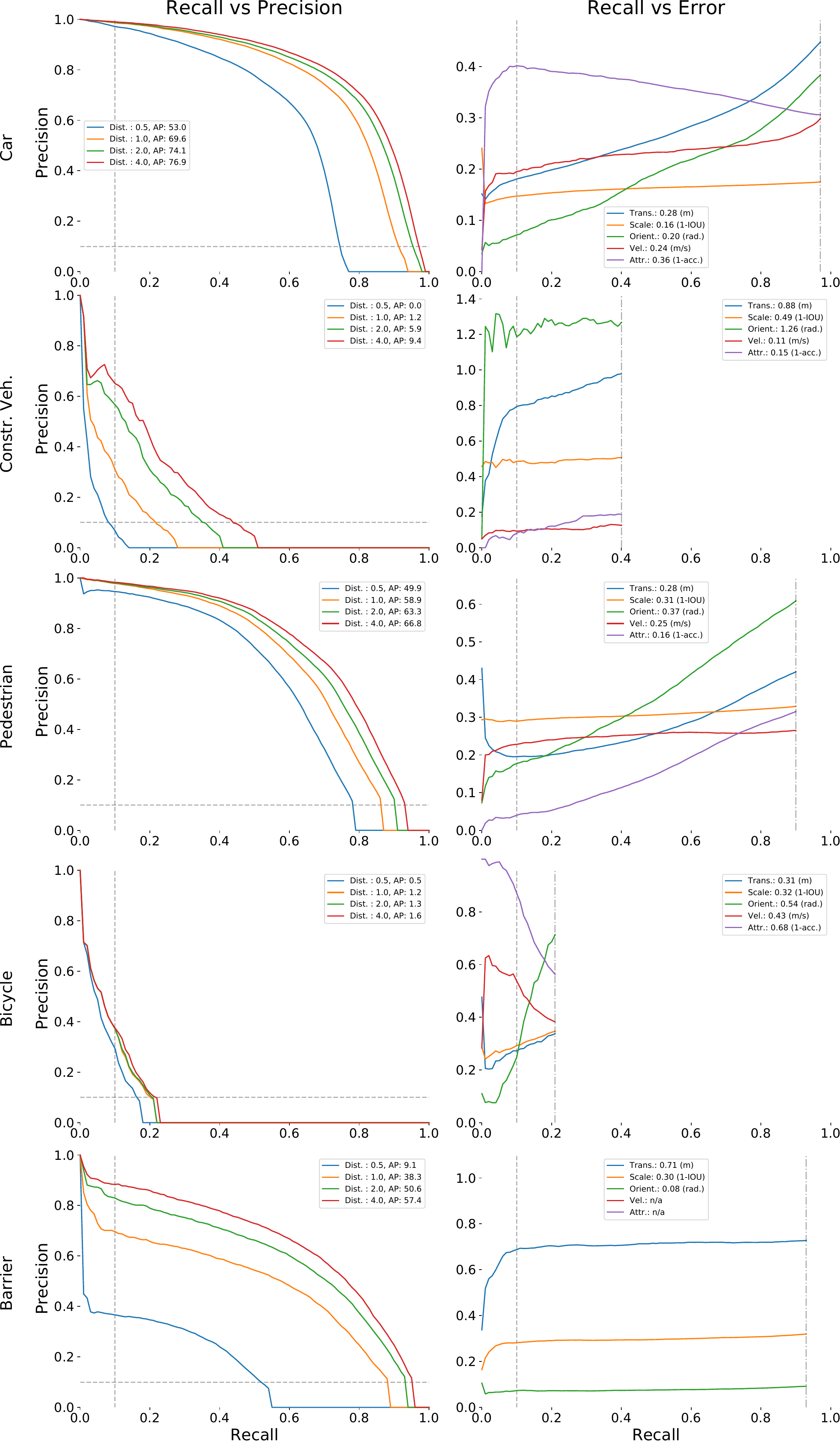}
\end{center}
\squeeze
\vspace{-2mm}
\caption{Per class results for PointPillars on the nuScenes test set taken  from the detection leaderboard.
}
\label{fig:detection-pp}
\end{figure}

\mypar{Per-class analysis.}
The per class performance of PointPillars~\cite{pointpillars} is shown in \tableref{table:class_results}-SM (top) and \figref{fig:detection-pp}-SM.
The network performed best overall on cars and pedestrians which are the two most common categories.
The worst performing categories were bicycles and construction vehicles, two of the rarest categories that also present additional challenges.
Construction vehicles pose a unique challenge due to their high variation in size and shape.
While the translational error is similar for cars and pedestrians, the orientation error for pedestrians (\ang{21}) is higher than that of cars (\ang{11}).
This smaller orientation error for cars is expected since cars have a greater distinction between their front and side profile relative to pedestrians.
The vehicle velocity estimates are promising (e.g. $0.24$ m/s AVE for the \emph{car} class) considering the typical speed of a vehicle in the city would be $10$ to $15$ m/s.

\begin{table}[]
\footnotesize
\begin{tabular}{| C{1.9cm} | c | c | c | c | c | c |} \hline
\multicolumn{7}{| c |}{\textbf{\small PointPillars}} \\
\hline
\textbf{Class}  &   \textbf{AP}  &   \textbf{ATE} &   \textbf{ASE} & \textbf{AOE}   & \textbf{AVE}   & \textbf{AAE}   \\ \hline \hline
Barrier         &    38.9   &   0.71  &   0.30  &   0.08  &    N/A  &    N/A  \\ \hline
Bicycle         &     1.1   &   0.31  &   0.32  &   0.54  &   0.43  &   0.68  \\ \hline
Bus             &    28.2   &   0.56  &   0.20  &   0.25  &   0.42  &   0.34  \\ \hline
Car             &    68.4   &   0.28  &   0.16  &   0.20  &   0.24  &   0.36  \\ \hline
Constr. Veh.    &     4.1   &   0.89  &   0.49  &   1.26  &   0.11  &   0.15  \\ \hline
Motorcycle      &    27.4   &   0.36  &   0.29  &   0.79  &   0.63  &   0.64  \\ \hline
Pedestrian      &    59.7   &   0.28  &   0.31  &   0.37  &   0.25  &   0.16  \\ \hline
Traffic Cone    &    30.8   &   0.40  &   0.39  &    N/A  &    N/A  &    N/A  \\ \hline
Trailer         &    23.4   &   0.89  &   0.20  &   0.83  &   0.20  &   0.21  \\ \hline
Truck           &    23.0   &   0.49  &   0.23  &   0.18  &   0.25  &   0.41  \\ \hline
\hline
\textbf{Mean}   &    30.5   &   0.52  &   0.29  &   0.50  &   0.32  &   0.37   \\ \hline
\end{tabular}
\begin{tabular}{| C{1.9cm} | c | c | c | c | c | c |} \hline
\multicolumn{7}{| c |}{\textbf{MonoDIS}} \\
\hline
\textbf{Class}  &   \textbf{AP}  &   \textbf{ATE} &   \textbf{ASE} & \textbf{AOE}   & \textbf{AVE}   & \textbf{AAE}   \\ \hline \hline
Barrier         &   51.1    &   0.53  &   0.29  &   0.15  &    N/A  &    N/A  \\ \hline
Bicycle         &   24.5    &   0.71  &   0.30  &   1.04  &   0.93  &   0.01  \\ \hline
Bus             &   18.8    &   0.84  &   0.19  &   0.12  &   2.86  &   0.30  \\ \hline
Car             &   47.8    &   0.61  &   0.15  &   0.07  &   1.78  &   0.12  \\ \hline
Constr. Veh.    &    7.4    &   1.03  &   0.39  &   0.89  &   0.38  &   0.15  \\ \hline
Motorcycle      &   29.0    &   0.66  &   0.24  &   0.51  &   3.15  &   0.02  \\ \hline
Pedestrian      &   37.0    &   0.70  &   0.31  &   1.27  &   0.89  &   0.18  \\ \hline
Traffic Cone    &   48.7    &   0.50  &   0.36  &    N/A  &    N/A  &    N/A  \\ \hline
Trailer         &   17.6    &   1.03  &   0.20  &   0.78  &   0.64  &   0.15  \\ \hline
Truck           &   22.0    &   0.78  &   0.20  &   0.08  &   1.80  &   0.14  \\ \hline
\hline
\textbf{Mean}   &   30.4    &   0.74  &   0.26  &   0.55  &   1.55  &   0.13  \\ \hline
\end{tabular}
\vspace{+1mm}
\caption{
Detailed detection performance for PointPillars~\cite{pointpillars} (top) and MonoDIS~\cite{monodis} (bottom) on the test set.
AP: average precision averaged over distance thresholds (\%),
ATE: average translation error (m),
ASE: average scale error (1-IOU),
AOE: average orientation error (rad),
AVE: average velocity error (m/s),
AAE: average attribute error ($1-acc.$),
N/A: not applicable (\secref{sec:tasks:detection}).
nuScenes Detection Score (NDS) = 45.3\% (PointPillars) and 38.4\% (MonoDIS).}
\label{table:class_results}
\end{table}

\begin{figure}
\begin{center}
\includegraphics[width=\linewidth]{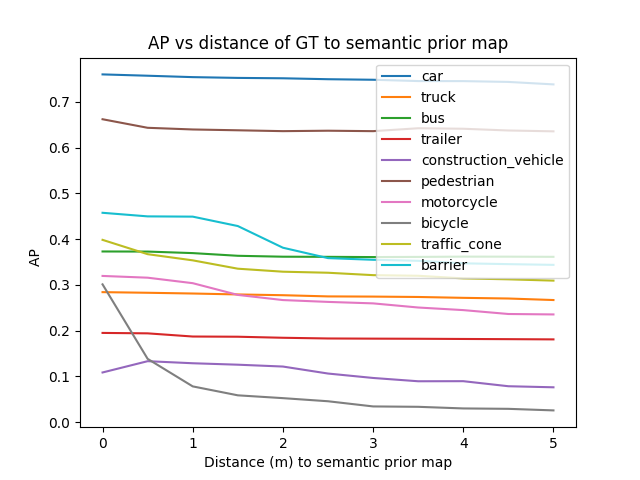}
\end{center}
\squeeze
\vspace{-3mm}
\caption{PointPillars~\cite{pointpillars} detection performance vs. semantic prior map location on the val set.
For the best lidar network (10 lidar sweeps with ImageNet pretraining), the predictions and ground truth annotations were only included if within a given distance of the semantic prior map.
}
\vspace{-2mm}
\label{fig:lidar_ab_vs_sp}
\end{figure}

\begin{table*}
\footnotesize
\begin{tabular}{| C{2cm} | C{1.7cm} | C{1.7cm} | C{1.7cm} | C{1.7cm} | C{1.7cm} | C{1.7cm} | C{1.7cm} |} \hline
\multirow{2}{*}{\textbf{Method}} & \textbf{sAMOTA} & \textbf{AMOTP} & \textbf{$\text{sMOTA}_r$} & \textbf{MOTA} & \textbf{MOTP} & \textbf{TID} & \textbf{LGD} \\ \cline{2-8}
\textbf & (\%) & (m) & (\%) & (\%) & (m) & (s) & (s) \\ \hline \hline
Stan~\cite{stanford2020nuscenes}    &   55.0    &   0.80    &   76.8  &   45.9  &   0.35  &   0.96  &   1.38    \\ \hline
VVte                                &   37.1    &   1.11    &   68.4  &   30.8  &   0.41  &   0.94  &   1.58    \\ \hline
Megvii~\cite{megvii}                &   15.1    &   1.50    &   55.2  &   15.4  &   0.40  &   1.97  &   3.74    \\ \hline
CeOp                                &   10.8    &   0.99    &   26.7  &    8.5  &   0.35  &   1.72  &   3.18    \\ \hline
CeVi$^{\dag}$                       &    4.6    &   1.54    &   23.1  &    4.3  &   0.75  &   2.06  &   3.82    \\ \hline
PP~\cite{pointpillars}              &    2.9    &   1.70    &   24.3  &    4.5  &   0.82  &   4.57  &   5.93    \\ \hline
MDIS~\cite{monodis}$^{\dag}$        &    1.8    &   1.79    &    9.1  &    2.0  &   0.90  &   1.41  &   3.35    \\ \hline
\end{tabular}
\vspace{+1mm}
\caption{
Tracking results on the test set of nuScenes.
PointPillars, MonoDIS (MaAB) and Megvii (MeAB) are submissions from the detection challenge, each using the AB3DMOT~\cite{ab3dmot} tracking baseline.
StanfordIPRL-TRI (Stan), VVte (VV-team), CenterTrack-Open (CeOp) and CenterTrack-Vision (CeVi) are the top submissions to the nuScenes tracking challenge leaderboard.
$({\dag})$ use only monocular camera images as input. CeOp uses lidar and camera. All other methods use only lidar.
}
\label{table:tracking_challenge}
\end{table*}

\mypar{Semantic map filtering.}
In \secref{sec:experiments:analysis} and \tableref{table:class_results}-SM we show that the PointPillars baseline achieves only an AP of $1\%$ on the \emph{bicycle} class.
However, when filtering both the predictions and ground truth to only include boxes on the semantic map prior\footnote{Defined here as the union of roads and sidewalks.}, the AP increases to $30\%$.
This observation can be seen in \figref{fig:lidar_ab_vs_sp}-SM, where we plot the AP at different distances of the ground truth to the semantic map prior.
As seen, the AP drops when the matched GT is farther from the semantic map prior.
Again, this is likely because bicycles away from the semantic map tend to be parked and occluded with low visibility.

\mypar{Tracking challenge results.}
In \tableref{table:tracking_challenge} we present the results of the 2019 nuScenes tracking challenge.
Stan~\cite{stanford2020nuscenes} use the Mahalanobis distance for matching, significantly outperforming the strongest baseline ($+40\%$ sAMOTA) and setting a new state-of-the-art on the nuScenes tracking benchmark.
As expected, the two methods using only monocular camera images perform poorly (CeVi and MDIS).
Similar to \secref{sec:experiments}, we observe that the metrics are highly correlated, with notable exceptions for MDIS LGD and CeOp AMOTP.
Note that all methods use a tracking-by-detection approach.
With the exception of CeOp and CeVi, all methods use a Kalman filter~\cite{kalmanfilter}.

\end{document}